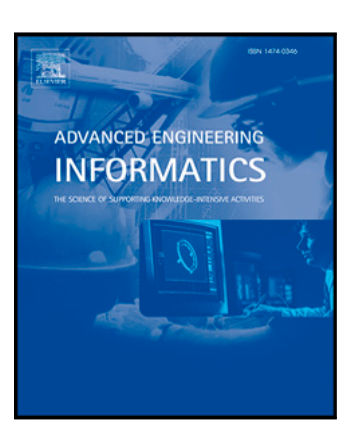

Full length article

# Harnessing human expertise for high-precision robotic assembly in industrialized construction: A sample-efficient installer-in-the-loop interactive reinforcement learning framework ☆

Zekai Jin, Huiguang Wang, Xiaoning Sun, Yi Shao *

*Department of Civil Engineering, McGill University, Montreal, Quebec, Canada*

ARTICLE INFO

ABSTRACT

Industrialized construction imposes stringent precision requirements on the robotic assembly of modular components such as prefabricated window units. In such tolerance-critical operations, the central bottleneck is not only mechanical clearance but also the difficulty of converting tacit installer expertise into data-efficient autonomy: sparse acceptance feedback, pronounced contact variability, and millimeter-scale geometric constraints jointly constitute a formidable sample-efficiency challenge. This study presents an installer-in-the-loop interactive reinforcement learning framework that acquires such expertise through three complementary channels: offline teleoperated demonstrations, sparse event-driven binary takeovers at contact-failure boundaries, and acceptance-aligned terminal rewards, all logged under a unified schema for traceable offline-to-online adaptation. A temporally abstract action-sequence policy built on Q-chunking with Flow Q-Learning captures multimodal recovery maneuvers under sparse terminal rewards, while a non-updating warm-start phase stabilizes the offline-to-online transition. The framework is evaluated in MuJoCo across the full operational workflow, from suction acquisition through clearance-limited seating, under both structured staging and end-to-end randomized placement. Within a defined stress-test regime characterized by 2 mm per-side clearance, bounded pose perturbations, and friction randomization, the proposed pipeline attains 100% autonomous seating with 12–15 min of cumulative installer supervision over 3.0 h of online training (12 min in Experiment A; 15 min in Experiment B), and reaches the 95% success milestone in approximately 0.5 h and 1.5 h, respectively. Beyond success rate, wall-clock adaptation time, cumulative takeover minutes, intervention-rate decay, and stage-wise failure attribution are reported to inform supervision budgeting. Ablations isolate the complementary contributions of temporal abstraction, installer intervention, and warm-start value calibration.

## 1. Introduction

Chronic productivity stagnation, escalating labor costs, and acute shortages of skilled tradespeople continue to beset the construction industry [1]. These structural pressures have catalyzed sustained interest in robotic automation of repetitive, physically demanding building-envelope operations [2]. Among candidate processes, prefabricated window and curtain wall installation represents a particularly consequential automation bottleneck: the workflow spans suction acquisition, transport, alignment, insertion, and final seating, and although the upstream stages broadly resemble coarse pick-and-place operations, overall system reliability is dominated by the terminal insertion–seating phase, where tight geometric clearance and sustained multi-surface contact are the principal sources of rework and autonomy breakdown.

This last-centimeter insertion regime, the terminal phase during which the unit transitions from pre-positioned alignment into sustained contact with the wall opening, demands that heavy and structurally fragile prefabricated units be seated with millimeter-level precision [3,4]. Under tight geometric constraints, insertion becomes contact-dominated, and even modest pose deviations can precipitate binding, wedging, or jamming [5]. Discrepancies between nominal BIM geometry and realized as-built conditions further invalidate pre-programmed execution routines [6,7]. Sustained multi-surface contact restricts the manifold of feasible motion and induces self-locking frictional effects; successful recovery consequently demands temporally extended, mode-dependent corrective sequences, including partial retraction, small-angle pivoting, and diagonal sliding, whose selection is governed by evolving contact history. Executing such maneuvers

☆ This article is part of a Special issue entitled: 'ADVEI_AI & Robotics for IC' published in Advanced Engineering Informatics.

* Corresponding author.
*E-mail address:* yi.shao2@mcgill.ca (Y. Shao).

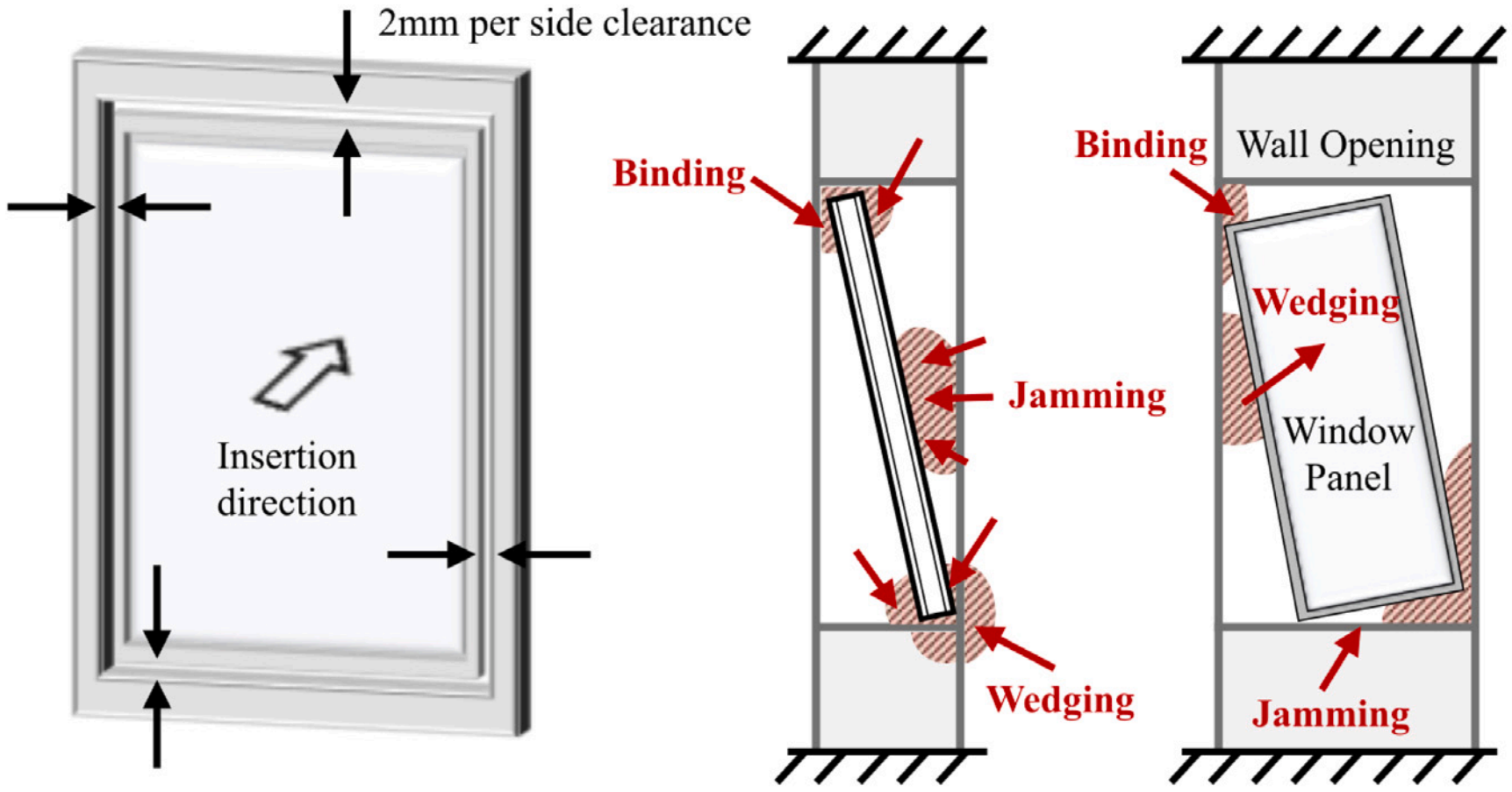

(a) Perspective insertion schematic. (b) Sectional failure modes: binding, wedging & jamming.

**Fig. 1.** Mechanics of terminal-phase failure modes in tolerance-critical window seating under millimeter-scale clearance. (a) Insertion schematic illustrating the transition into a wall opening with 2 mm per-side clearance. (Front-oriented schematic; mild perspective, not an orthographic elevation.) (b) Contact-induced failure mechanisms during insertion and seating. (Left: lateral cross-section; right: front-view schematic of the wall opening.) Taxonomy of *binding* (motion arrest attributable to distributed frictional contact), *wedging* (geometric self-locking under normal loading induced by misalignment), and *jamming* (persistent multi-surface contact necessitating non-monotonic corrective maneuvers). These modes delineate the high-risk boundary regimes where installer expertise is formalized to curtail autonomy breakdowns by precluding entry into low-recoverability contact states.

is fundamentally an expertise-dependent engineering problem: reliable seating hinges on tacit procedural knowledge of when to retract, how far to pivot, and under what contact conditions to reinsert, knowledge that admits neither analytical specification nor derivation from nominal BIM geometry alone. Formalizing and operationalizing this knowledge for robust autonomy constitutes the central engineering informatics challenge of the present study. Fig. 1 summarizes the mechanics of this clearance-limited contact regime and defines the failure-mode taxonomy used throughout.

Existing construction automation systems perform well at large-scale material handling, perception-driven planning, and free-space alignment, yet robustness degrades sharply once prolonged contact is established during insertion and seating [8,9]. Skilled installers resolve such deadlocks through compact corrective sequences executed under latent contact modes, whereas scripted controllers routinely fail when recovery requires coordinated multi-step behavior under sparse success criteria [10,11]. Learning-based policies can acquire contact-rich recovery behaviors under acceptance-aligned evaluation objectives. The present framework captures installer knowledge through teleoperated demonstrations, event-driven binary takeovers, and terminal acceptance signals, all recorded under a unified schema that enables traceable offline-to-online adaptation (Sections 2 and 3.2).

The learning core builds on *Q-chunking with Flow Q-Learning* (QC-FQL) [12], adapted to the installer-in-the-loop data regime under sparse, acceptance-aligned seating feedback within a clearance-limited façade workflow; Section 3.5 details the flow teacher–student realization and chunk-level value learning.

Reinforcement learning (RL) enables closed-loop policy acquisition without explicit contact models [13–15], yet sparse rewards and extended temporal horizons render tolerance-critical seating particularly challenging [16–20]. Undirected trial-and-error incurs prohibitive cost under contact-rich dynamics, while dense reward shaping can systematically misalign with acceptance-based seating criteria. The majority of construction-oriented RL studies either exclude domain experts from the learning loop or focus on lower-contact operational stages [21–24]. This study therefore operationalizes an installer-in-the-loop setting in which installers supply offline demonstrations and sparse event-driven interventions (Fig. 2) [19,25,26]. Simulation is employed to probe failure modes safely; the limits of transfer to physical settings are discussed in Section 5.

Temporally extended action sequences with a generative flow-based prior (QC-FQL; introduced above) and a non-updating warm-start phase for stable offline-to-online transfer are adopted throughout; Fig. 3 summarizes the problem setting and the three training stages.

Rather than introducing a new generic RL algorithm, the methodological contribution lies in a domain-specific engineering informatics instantiation of QC-FQL for chunk-level value learning and HIL-SERL-style binary takeovers for human guidance [12,26]. This instantiation combines a unified logging schema for installer knowledge artifacts, acceptance-aligned terminal rewards, and simulation-based deployment-oriented metrics, including wall-clock adaptation time, takeover minutes, and stage-wise failure attribution, for tolerance-critical façade assembly. By decomposing high-precision construction skill acquisition into structured demonstrations, event-triggered recovery interventions, and acceptance-based evaluation, the framework provides a task-grounded methodological template for fine-grained robotic skills in construction. Full technical details appear in Sections 3.2, 3.5 and 4.2.

To the authors' knowledge, this study is among the first systematic investigations to unify installer-guided RL for tolerance-critical construction assembly under acceptance-aligned evaluation. Wall-clock adaptation time and intervention burden are quantified, and ablations are reported that isolate the contribution of each major protocol component. The study is aligned with the human-centered automation ethos of Construction 5.0 [23,27] and is evaluated in MuJoCo under a simulated stress-test regime [28].

Accordingly, this study makes four principal contributions:

- A MuJoCo benchmark is formulated for end-to-end prefabricated window installation, in which upstream variability propagates into clearance-limited insertion and seating under 2 mm per-side tolerance, multi-surface friction, and emergent binding, wedging, and jamming.
- A three-stage installer-in-the-loop learning protocol is developed, integrating offline demonstrations, non-updating warm-start calibration, and online fine-tuning with sparse event-driven takeovers under a unified logging schema for traceable supervision accounting.
- A temporally abstract action-chunk policy with a flow-based multimodal prior is instantiated to represent recovery maneuvers

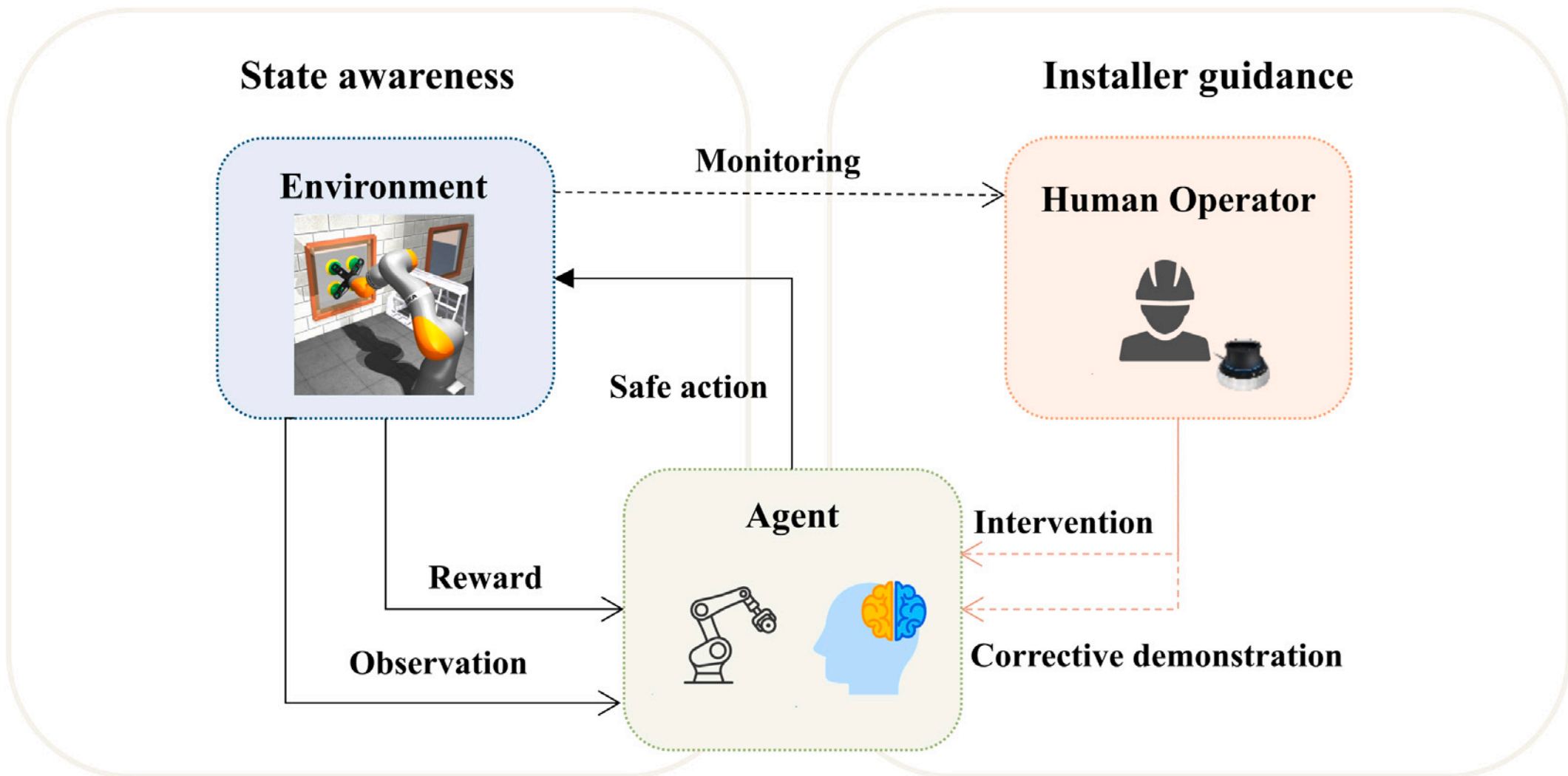

**Fig. 2.** Schematic of the proposed installer-in-the-loop interactive RL interface. For interpretability, the layout groups *state-awareness* elements (observation streams and monitoring) on one side and *installer-to-agent guidance* elements (intervention/takeover and corrective demonstrations) on the other; arrows indicate information and control flow. This arrangement separates state-awareness channels from installer-guidance channels and avoids implying spurious coupling between observation and intervention.

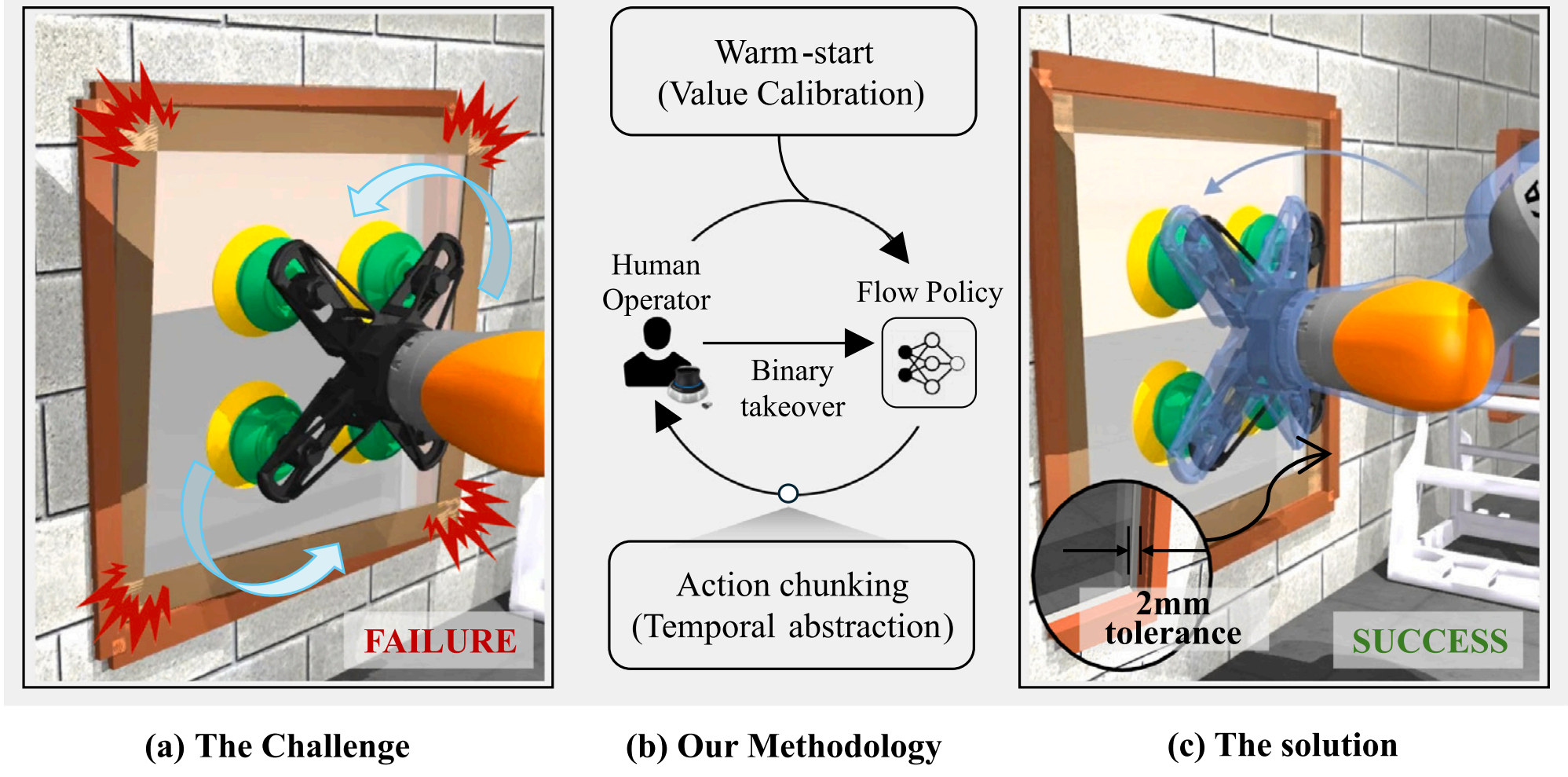

**Fig. 3.** Roadmap of the proposed installer-in-the-loop interactive RL paradigm for resolving contact-rich assembly deadlocks. (a) Failure regime under millimeter-scale clearance, characterized by complex frictional contact and inherently multimodal recovery paths. (b) Three-stage training paradigm: offline behavioral initialization, non-updating warm-start value calibration, and online fine-tuning with event-driven binary takeovers. (c) Target behavior: temporally abstract action-sequence execution enabling coherent multi-step recovery and reliable seating under 2 mm per-side tolerance.

such as retreat–pivot–reinsert, enabling chunk-level credit assignment under sparse terminal rewards; ablations indicate that both temporal abstraction and multimodal sequence modeling are necessary within this stress-test regime.

- A deployment-oriented evaluation protocol is introduced beyond scalar success rate, reporting wall-clock adaptation time, takeover minutes, intervention-rate decay, and stage-wise failure attribution under matched-budget comparisons to support supervision budgeting and failure diagnosis in simulation, without constituting field guarantees.

## 2. Related work

The study is situated across three technical foundations: building-envelope automation, RL-based control for construction robotics, and interactive human-in-the-loop RL for contact-rich manipulation.

### 2.1. Building-envelope installation automation

Robotic automation for building-envelope installation has progressed from bulk material handling toward increasingly autonomous placement of prefabricated components [2,8]. In practice, system capability is strongly conditioned by operational scale and control architecture, which together determine the extent of geometric variability, frictional contact, and clearance-critical seating that can be accommodated without site-specific tuning. At the large-scale end, task-specialized platforms, including gantry-like façade systems and cable-driven parallel robots for curtain wall assembly, are designed to deliver the payload capacity and reach required for transport and coarse placement across expansive workspaces [9,29]. Such platforms provide mechanical leverage, yet local precision during late-stage seating is often compromised by structural compliance, residual oscillation, and uncertainty in sensing and pose estimation. A well-documented accuracy–workspace trade-off emerges: increased reach and payload capacity tend to erode local positional fidelity [30]. Consequently, insertion and final seating under millimeter-scale clearances and sustained asymmetric contact routinely remain dependent on manual intervention or extensive on-site calibration [2,10].

A complementary research strand deploys general-purpose manipulators augmented with perception, localization, and motion-planning capabilities [10,31]. These systems offer flexibility across diverse site layouts and component geometries and can execute free-space alignment and obstacle avoidance effectively; however, robustness tends to degrade once prolonged contact is established [10]. Under non-uniform friction and as-built tolerances, modest pose errors can escalate into wedging or jamming during seating. Although impedance and force-feedback control can regulate interaction forces, reliable window installation frequently demands multi-step corrective sequences, including partial retraction, diagonal sliding, and small-angle pivoting, whose selection is conditioned on contact history in ways that resist enumeration as a small set of discrete modes [3,11,32].

From a workflow-control perspective, window installation differs from classical industrial peg-in-hole insertion in two important respects. First, construction openings exhibit as-built geometric deviations and partial observability during seating [33]. Second, recovery from contact deadlocks is long-horizon and history-dependent: successful seating typically requires temporally extended corrective sequences rather than brief terminal routines. These properties motivate learning-based closed-loop policies that operate inside, rather than around, the contact-rich regime.

From an engineering informatics standpoint, however, the bottleneck is not solely one of controller design but of knowledge availability: how the procedural expertise of skilled installers, which governs when to retract, how far to pivot, and under what contact conditions to reinsert, is formalized, acquired, and updated during commissioning. Construction practice is fundamentally knowledge-intensive, yet ontology-based and case-based reasoning frameworks predominantly formalize *what* to build and *where* to place components, not *how* operators physically resolve contact-induced failures [34,35]. Data-driven approaches frequently treat installer knowledge as a static offline asset and provide no mechanism for deployment-time refinement [23]. The present study addresses this gap by formalizing demonstrations, interventions, and acceptance signals as typed knowledge artifacts that are progressively integrated into the policy during commissioning, thereby operationalizing tacit execution knowledge as a traceable engineering informatics process rather than an algorithmic training convenience.

### *2.2. RL in construction robotics*

In construction robotics, RL has been investigated as a route to contact-aware feedback control under uncertain site conditions, particularly when accurate analytical contact models are unavailable. Representative examples span hierarchical RL for window panel installation [24], tolerance-sensitive timber assembly [36], heavy equipment interaction such as soil–tool dynamics [37], and geometry-driven design and assembly planning for prefabricated workflows [38,39]. More recently, learning-based systems have been directed toward contact-sensitive building-envelope workflows including robotic glass installation [31]. Despite this progress, many construction-RL formulations reduce exposure to sustained multi-contact recovery: tasks are frequently short-horizon, quasi-static, or framed as single-stage insertions, limiting the need for long-horizon corrective behavior under jamming conditions [36,37]. Credit assignment under sparse success signals remains a fundamental bottleneck, as step-wise methods such as PPO [40] and SAC [41] propagate value locally in time, and exploration readily stalls when success requires coordinated behavior sustained over extended horizons [42–45]. In clearance-limited insertion, this manifests as persistent failure to discover and reliably execute multi-step recovery sequences, such as retreat, pivot, and reinsert, near contact boundaries.

Temporal abstraction via action chunking provides a complementary interface by assigning value to and selecting among short action sequences rather than individual commands [12,46]. In robotics, temporal abstraction has been studied through motion primitives and task-frame formulations [47–49], and recent work predicts short action sequences directly from state observations. In construction-oriented learning, action chunking has been validated primarily under imitation learning and short-horizon settings [46,50], whereas TD-based evidence suggests benefits for sparse-reward long-horizon problems through chunk-level value propagation [12,51]. Existing RL approaches in construction still underrepresent long-horizon recovery under sparse rewards. The present chunked action framework addresses this gap by combining chunk-level value propagation with a multimodal generative prior in an end-to-end contact-rich workflow. This combination remains underexplored in construction-oriented RL and is particularly consequential for tolerance-critical assembly, where recovery is both temporally extended and intrinsically multimodal.

### *2.3. Operator guidance: Interactive human-in-the-loop RL*

The integration of human expertise has long been recognized as a central mechanism for accelerating robotic policy learning. Early foundations in interactive imitation learning, most notably the DAgger framework [52], established the principle of collecting expert corrections on the induced state distribution of the learner to mitigate compounding errors and distribution shift. This principle evolved into intervention-based variants such as HG-DAgger [53], which replace continuous labeling with event-driven takeovers. While such methods deliver targeted data around imminent failures, pure imitation learning remains bounded by the demonstrator's performance ceiling and does not explicitly optimize for task success beyond behavioral mimicry [22,54]. From a learning-theoretic standpoint, event-driven takeovers bias data collection toward high-consequence contact-boundary states, where compounding errors would otherwise drive the system into low-recoverability regimes. Interpreted as safety-aware boundary constraints, this sampling shift provides disproportionately informative recovery experience and improves sample efficiency in tolerance-critical assembly [26].

Building on this grounding, interactive human-in-the-loop RL incorporates interventions directly into off-policy RL to optimize beyond the demonstrator's performance ceiling. The primary motivation is to counter exploration collapse in sparse-reward, contact-rich environments, where successful terminal states occupy a narrow region under millimeter-scale clearance. Following the design principles of HIL-SERL [26], binary takeovers serve a coherent dual role: they enforce a safety–recoverability envelope near high-risk contact boundaries while concentrating replay data on boundary states and recovery segments rarely encountered through unguided exploration. This targeted redistribution of the visitation distribution can improve sample efficiency under sparse rewards without dense reward shaping or hand-crafted recovery logic. Operationally, it corresponds to variable autonomy via control sharing, in which full autonomy is maintained over routine execution phases and reduced when an expert judges that continued autonomous operation is unlikely to remain within the recoverable region [18,55].

Beyond exploration, offline-to-online fine-tuning introduces stability challenges under distribution shift, particularly when discontinuous contact impulses and contact-mode transitions are underrepresented in the demonstration corpus. As highlighted in HIL-SERL [26], concentrating interventions near boundary regimes supplies high-value anchors that stabilize critic learning during online adaptation, and brief warm-start data-collection phases can recalibrate the critic to the online interaction distribution before gradient updates begin [56,57]. Prior human-in-the-loop RL formulations typically retain step-wise control and rely on unimodal policies, which may struggle to resolve contact deadlocks that exhibit intrinsic multimodality in the last-millimeter regime. The present study extends the installer-in-the-loop RL paradigm by integrating temporal abstraction with a multimodal sequence prior, enabling the acquisition of temporally coherent and multimodal recovery maneuvers under sparse terminal rewards.

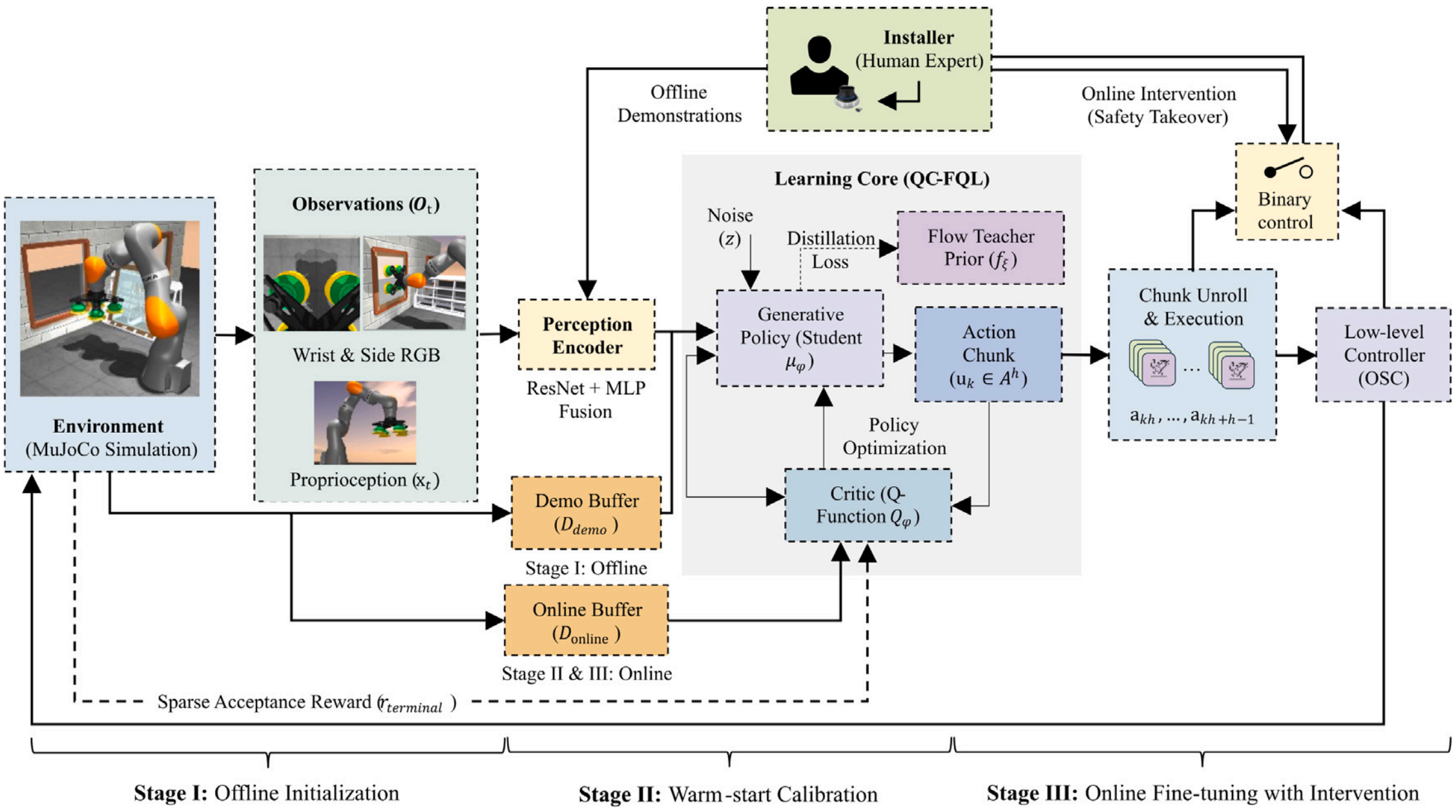

**Fig. 4.** Proposed installer-in-the-loop RL pipeline. Demonstrations and online interaction data are collected under a shared chunked action interface and stored in $\mathcal{D}_{\text{demo}}$ and $\mathcal{D}_{\text{online}}$, respectively. The QC-FQL core performs chunk-level value learning via a teacher–student factorization: a flow-based behavior model $f_\xi$ provides multimodal sequence priors and distillation targets for a noise-conditioned student actor $\mu_\psi$, while the critic $Q_\phi$ evaluates executed action chunks. During online training, event-driven binary takeovers may override the policy at the high-level action interface; executed chunks are unrolled for control execution under the same low-level tracking controller employed throughout.

## 3. Methodology

The methodology integrates installer expertise into a temporally abstract RL core for sample-efficient, high-precision window installation. The presentation first provides a high-level overview and then defines the learning environment, the intervention protocol, and the optimization objectives.

### 3.1. Overview of the proposed framework

From a construction-automation perspective, the workflow is partitioned so that the robot executes routine segments autonomously while the installer intervenes only when the interaction approaches tolerance boundaries at which safe and recoverable completion within the episode horizon becomes uncertain. The policy is initialized from teleoperated demonstrations, followed by a brief non-updating warm-start phase for value calibration, and subsequently fine-tuned online with event-driven binary takeovers integrated into the QC-FQL learning core. Fig. 4 summarizes the proposed installer-in-the-loop RL pipeline and illustrates how demonstrations, warm-start calibration, and event-driven takeovers interact under a shared chunked control interface. This shared interface also serves as a data-logging boundary, ensuring that demonstrations, interventions, and autonomous execution are recorded under a unified schema for traceable offline-to-online adaptation.

The learning core follows QC-FQL [12], which performs chunk-level Q-learning in a fixed-horizon action-sequence space and distills a flow-based multimodal behavior prior into a constant-time actor for real-time execution. The framework operates on fixed-horizon action chunks rather than step-wise commands, so each decision corresponds to a short motion segment consistent with practical corrective maneuvers such as pivot–slide–reinsert. This temporal abstraction supports chunk-level value learning under terminal acceptance feedback and represents multimodal recovery without action blending. The chunked decision interface is defined in Section 3.2, the MuJoCo benchmark with contact randomization in Section 3.3, and the installer takeover protocol in Section 3.4. The chunk-level learning objective and system implementation are presented in Sections 3.5 and 3.6, respectively.

### 3.2. Preliminaries and problem formulation

*Task objective and finite-horizon mdp.* Long-horizon window installation is formulated as an offline-to-online RL problem with sparse, acceptance-aligned success. A policy $\pi$ is initialized from a fixed teleoperated demonstration dataset $\mathcal{D}_{\text{demo}}$ and subsequently refined online with additional interaction data $\mathcal{D}_{\text{online}}$ collected under an identical sensing and control interface. The interaction is modeled as a finite-horizon Markov Decision Process (MDP) $\mathcal{M} = (\mathcal{S}, \mathcal{A}, P, r, \rho, H, \gamma)$, where $\mathcal{S}$ denotes the space of state observations available to the agent, $\mathcal{A}$ the action space, $P$ the environment transition kernel induced by robot dynamics and contact interactions, $\rho$ the initial state distribution, $H$ the episode horizon, and $\gamma \in (0,1)$ the discount factor. In practice, the agent does not access the full latent physical state (e.g., contact modes or frictional constraints) but instead receives sensor-derived observations from onboard perception and proprioception. Following standard practice in vision-based robotics RL (e.g., HIL-SERL [26]), these observations serve as the learning-state input, and $s_t \in \mathcal{S}$ denotes the observable state at time step $t$. For notational convenience, $s_t \doteq o_t$ is used throughout.

*Sparse acceptance-aligned reward.* The reward is defined as a sparse terminal success signal, reflecting the binary acceptance-evaluation criteria of construction practice. Specifically:

$$r_t = \begin{cases} 1, & \text{if the window reaches acceptance-aligned seating at step } t, \\ 0, & \text{otherwise,} \end{cases} \tag{1}$$

which avoids proxy shaping and is directly aligned with construction acceptance semantics. Episodes terminate upon first satisfaction of the seating condition; otherwise, they terminate at the fixed horizon $H$. When termination occurs within a chunk, rewards after termination are zero and bootstrap terms are masked accordingly.

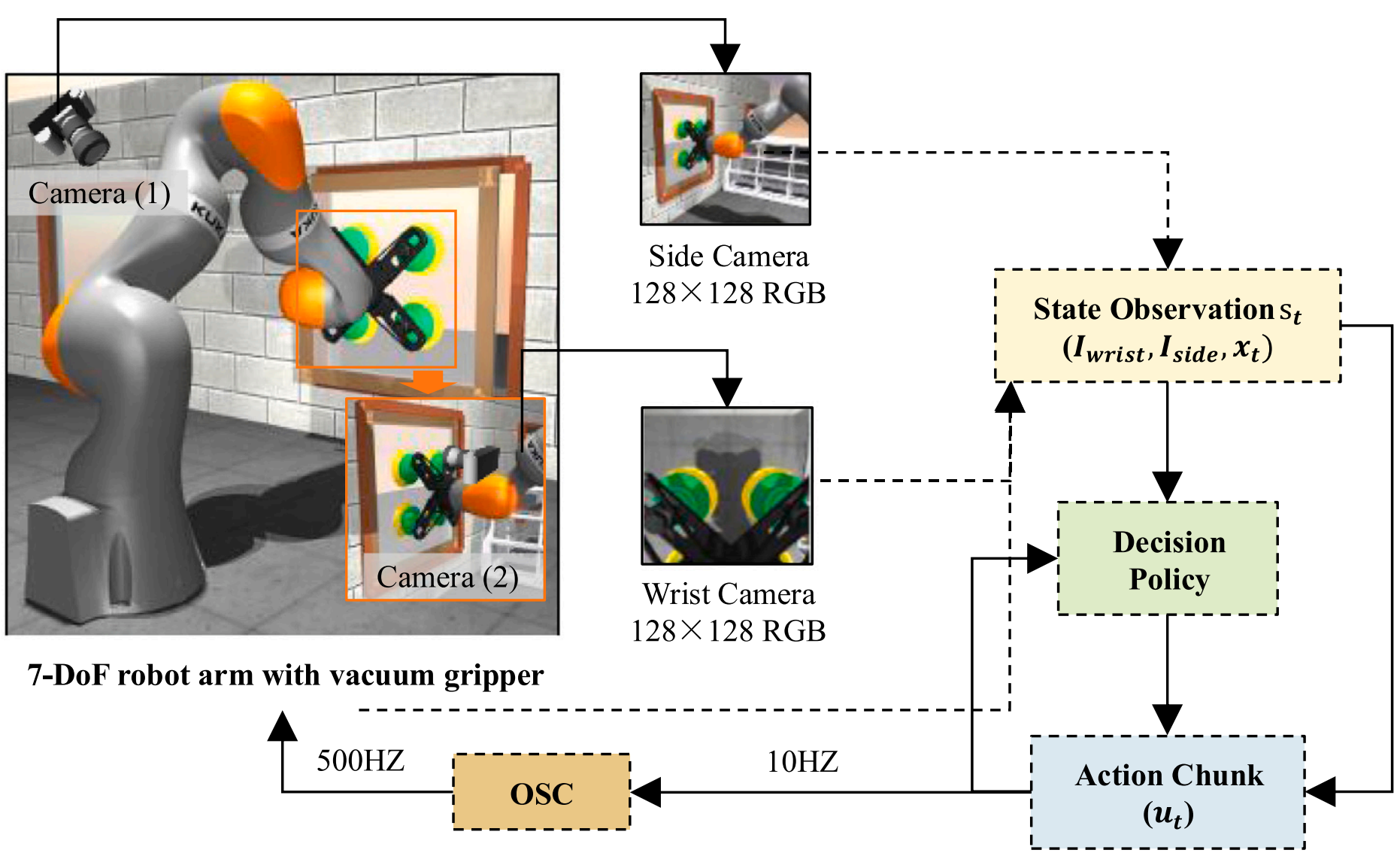

**Fig. 5.** System architecture for the proposed window installation framework. Dual-view RGB and proprioception constitute the state observation $s_t$. The environment step follows the 10 Hz high-level interface defined in Section 3.2.

*Structured knowledge artifacts.* The contribution of each interaction modality and its representation under the shared 10 Hz interface and replay buffers are stated below. *Demonstration artifacts* are episodes in the fixed offline dataset $\mathcal{D}_{\text{demo}}$, stored as step transitions $(s_t, a_t, r_t, s_{t+1})$ under the same observation–action schema as online training, including the chunked high-level interface; they initialize offline behavioral priors for the actor–critic stack. *Intervention artifacts* are transitions produced when binary takeovers replace the policy action during online training. Executed actions $a_t$, timestamps, and episode identifiers are logged exclusively to $\mathcal{D}_{\text{online}}$ together with the associated transitions, and chunk indices remain aligned as described in the paragraph *Logging and chunk-transition construction under takeovers* below. *Acceptance artifacts* are the sparse terminal reward signals in Eq. (1), with $r_t = 1$ only when seating satisfies the acceptance criteria and $r_t = 0$ otherwise, anchoring learning to construction quality-control semantics rather than dense shaping. During online fine-tuning, training batches are drawn from the union $\mathcal{D}_{\text{demo}} \cup \mathcal{D}_{\text{online}}$ at a fixed 1 : 1 ratio between offline demonstrations and online experience (Stage III, Section 3.4), rather than from a separate prioritized replay over artifact types. Together, these structured, typed knowledge artifacts supply complementary supervision: demonstrations initialize behavioral priors, interventions provide corrective boundary transitions, and acceptance artifacts anchor learning to task-level success semantics. The standardized logging and acceptance semantics further enable traceable knowledge reuse across future façade configurations.

*Observation and learning-state representation.* At each environment step $t$, the agent receives a state observation $s_t = (I_t^{\text{wrist}}, I_t^{\text{side}}, x_t) \in \mathcal{S}$. The environment operates at 10 Hz, and at each step an action $a_t$ is issued either by the policy or by the installer during intervention. Commanded targets are executed by a 500 Hz operational-space PD tracking controller (OSC) exhibiting impedance-like spring–damper behavior, providing the physical compliance required for contact-rich assembly. The OSC maps task-space pose errors to joint torques via operational-space inertia and Jacobian-based torque mapping, with gravity compensation and an additional nullspace joint PD term. Fig. 5 illustrates the overall system architecture and data flow. Here $I_t^{\text{wrist}}, I_t^{\text{side}} \in \mathbb{R}^{128\times128\times3}$ are RGB images from a wrist-mounted camera and a side camera, respectively, and $x_t \in \mathbb{R}^{18}$ is a proprioceptive vector encoding joint positions $q_t \in \mathbb{R}^7$, joint velocities $\dot{q}_t \in \mathbb{R}^7$, end-effector Cartesian position $p_t^{\text{ee}} \in \mathbb{R}^3$, and vacuum gripper status $g_t \in \mathbb{R}^1$.

In this interface, the agent observes realized robot motion (e.g., $p_t^{\text{ee}}$) while issuing incremental pose targets through the same action channel. This state representation supports contact-aware manipulation without requiring explicit force/torque sensing, maintaining compatibility with deployments in which only vision and robot proprioception are available. Under tracking control, contact constraints appear as reduced realized motion relative to commanded increments, together with characteristic changes in short-horizon transitions; the tracking outcome is therefore treated as a deployment-relevant contact proxy, following HIL-SERL [26]. Although contact dynamics are partially observable (e.g., latent contact modes and frictional constraints), an observation-based MDP formulation, widely adopted in vision-based robotic control [26,58], is employed here. Fixed-horizon chunking implicitly embeds short-term action–observation history into the decision unit, partially mitigating partial observability without recurrent policy architectures. Both the student actor and critic are feedforward; temporal coherence is induced instead by fixed-horizon chunking and a multimodal action-sequence prior (Section 3.5).

*Action interface and hierarchical execution.* The action space is $\mathcal{A} = \mathbb{R}^6 \times \{0, 1\}$ ($d_{\text{action}} = 7$), where each action $a_t = [\Delta p_t, \Delta r_t, g_t]$ issued at the 10 Hz environment step comprises a 6D incremental end-effector pose command $[\Delta p_t, \Delta r_t] \in \mathbb{R}^6$ (3D position $\Delta p_t$ and 3D orientation $\Delta r_t$, both expressed in the current end-effector frame) and a binary vacuum gripper command $g_t \in \{0, 1\}$. Pose increments are normalized to the unit hypercube $[-1, 1]^6$ via action scaling, while the vacuum command encodes discrete engagement states.

*Fixed-horizon action chunking.* To improve long-horizon credit assignment under the sparse terminal reward, fixed-length action chunking is adopted [12]. Let $h$ denote the chunk horizon (in environment steps). Unless stated otherwise, $h = 10$ is used throughout. At 10 Hz, $h = 10$ corresponds to a 1.0 s decision window. Following the Q-chunking perspective of Li et al. [12], operating RL in a chunked action space promotes temporally coherent high-level behavior and aligns TD backups with executed action segments. These properties are especially valuable under sparse terminal rewards; Li et al. further note that the optimal $h$ is task-dependent rather than universal. This timescale is consistent with teleoperation delay studies up to 1 s [59] and with observations of corrective maneuver timing during teleoperation. Accordingly, $h = 10$ is adopted as the **engineering default** for the headline experiments in Sections 4.4–4.5, where learning curves, baselines, and installer-in-the-loop variants are all reported at this horizon. Section 4.6 and Tables 5 and 6 present compact horizon scans for *Ours (w/o intervention)* to document how $h$ modulates final performance and, in Experiment B, stage-wise failure trade-offs. These tables are not used to assert strict superiority of any single $h$ on every metric, but to establish that chunking remains informative relative to step-wise control and that horizon choice interacts with staged versus end-to-end protocols.

At chunk index $k$ (corresponding to environment time $t = kh$), the high-level policy outputs a length-$h$ action sequence:

$$\mathbf{u}_k \doteq (a_{kh}, a_{kh+1}, \dots, a_{kh+h-1}) \in \mathcal{A}^h, \tag{2}$$

which is executed sequentially for $h$ environment steps unless interrupted by an installer takeover (Section 3.4). The accumulated discounted chunk return and effective discount factor are

$$R_k^{(h)} = \sum_{i=0}^{h-1} \gamma^i r_{kh+i}, \qquad \Gamma = \gamma^h. \tag{3}$$

*Logging and chunk-transition construction under takeovers.* During rollouts, the policy proposes a new chunk only when the action queue is exhausted, which ordinarily occurs at fixed chunk boundaries when $t$ is a multiple of $h$. If an installer takeover is initiated mid-chunk, the installer replaces the policy action at the shared 10 Hz action interface for as many steps as required. Although this resets the queue by clearing any remaining queued actions, chunk-transition construction remains indexed by fixed time steps, preserving temporal alignment. Concretely, the current chunk still ends at $t = kh+h$, and the next chunk is proposed from the updated observation at that fixed boundary. If a takeover spans multiple chunk boundaries, the installer continues issuing step-level actions, and the policy resumes at the first boundary following the conclusion of the takeover. This convention keeps the chunk-level learning framework internally consistent and integrates all executed actions, whether policy-generated or installer-supplied, into the same learning process.

Interaction is recorded at the 10 Hz environment step as $(s_t, a_t, r_t, s_{t+1})$, where $a_t$ denotes the executed action regardless of provenance. All learning quantities, including replay entries, Bellman backups, and intervention-rate statistics, are defined on this 10 Hz step index; the 500 Hz OSC is used solely for tracking and for converting takeover timesteps into wall-clock duration. Chunked transitions are constructed directly from the step log:

$$\big(s_{kh}, \mathbf{u}_k, R_k^{(h)}, s_{kh+h}\big), \tag{4}$$

where $\mathbf{u}_k$ is the length-$h$ executed action sequence over the next $h$ steps, regardless of action provenance. From the perspective of the induced SMDP, any executed sequence, whether generated by the policy or produced during a takeover, constitutes a valid macro-action. Value learning is therefore defined over realized action chunks rather than policy-proposed chunks. This keeps the critic grounded in executable recovery behaviors and preserves target–input alignment even under human intervention. The chunk return $R_k^{(h)}$ is computed from the rewards actually received during execution of this sequence, ensuring that value learning reflects the outcome of the executed actions.

### 3.3. Simulation environment setup

A MuJoCo benchmark for end-to-end window installation is implemented, encompassing the complete operational workflow from suction acquisition to final seating. Task progress in this regime depends sensitively on multi-surface friction and contact-mode transitions. Because these interactions are history-dependent and only partially observable, visually similar states can correspond to distinct contact configurations requiring qualitatively different recovery sequences [11, 44]. To expose sensitivity to construction variability, initial window pose is randomized within bounded tolerances, surface friction coefficients are sampled from prescribed ranges, and selected numerical contact parameters are perturbed within stable simulation regimes [60–62]. Unlike environments focused purely on transport, this benchmark treats tolerance-critical insertion as the primary bottleneck, reflecting the fact that upstream variation in grasp and alignment frequently propagates into late-stage contact failures. Fig. 6 visualizes the sequential milestones of both experimental settings, detailed further in Section 4.1.

The simulated workcell represents a façade installation environment comprising a prefabricated window unit, a wall opening lined with a wood buck, and a collaborative industrial manipulator equipped with a Robotiq PowerPick20 vacuum gripper. The simulated manipulator conforms to the kinematic and dynamic limits of a 7-DoF KUKA LBR iiwa 14 platform (payload 14 kg, reach 820 mm), with joint limits specified in Appendix. The full collision geometry of the gripper is modeled explicitly rather than approximated as a point constraint [24,63]. Vacuum engagement is implemented through a grasp-validity check evaluated at each 10 Hz environment step. When $g_t = 1$ and the validity condition is satisfied, suction is engaged and remains latched unless the policy issues $g_t = 0$ or the validity check fails. Suction is therefore not modeled as an always-on rigid constraint, and detailed suction dynamics are abstracted at the workflow level. Because the gripper occupies non-negligible volume near the opening, it constrains approach angles, insertion depth, and recovery maneuvers under tight clearance. The window is modeled as a rigid $0.5\,\mathrm{m} \times 0.5\,\mathrm{m}$ panel with explicit collision geometry. Limited substrate compliance is captured via MuJoCo compliant contact settings, with elevated friction coefficients on wood-buck surfaces to induce a friction-dominated insertion regime.

Success is defined as reaching a predefined seated target pose satisfying three acceptance-aligned criteria: (i) the Euclidean distance between the window geometric center and that of the predefined seated target at the specified insertion depth is less than $2\,\mathrm{mm}$; (ii) the cosine similarity between unit normals satisfies $\langle n_{\mathrm{win}}, n_{\mathrm{wall}} \rangle > 0.95$; and (iii) the window velocity magnitude is less than $1\,\mathrm{cm/s}$, ensuring stable seating rather than transient contact. These criteria enforce acceptance-level seating quality independently of the clearance definition, which governs geometric feasibility and the emergence of contact-induced failure modes.

In physical deployment, these criteria would correspond to verifiable acceptance checks such as vision-based pose estimation relative to a predefined reference frame, perimeter gap inspection, and residual motion verification via robot state estimation. Although the simulator provides noise-free state measurements, the criteria are designed to reflect practical seating verification procedures rather than task-specific shaping signals.

The simulator supports two operating modes under a shared observation–action interface. In teleoperation mode, an installer employs a 6-DoF SpaceMouse to collect demonstrations and to issue event-driven takeovers. In training mode, the policy executes autonomously. Both modes share identical observation streams and the chunked action interface defined in Section 3.2, ensuring consistency among demonstrations, interventions, and autonomous rollouts.

### 3.4. Installer guidance: Human-in-the-loop RL

Tabula-rasa learning is impractical in this setting, even in simulation [64]. The success signal is observed only at episode termination, informative contact-boundary states are rarely visited by unguided exploration, and undirected rollouts rapidly enter wedging or jamming configurations that are unlikely to recover within the episode horizon [32,44]. An installer-guided offline-to-online pipeline is therefore adopted, combining teleoperated demonstrations with sparse, event-driven interventions during online fine-tuning. The learner is off-policy and trains on a mixed replay buffer formed by a fixed demonstration dataset $\mathcal{D}_{\mathrm{demo}}$ and an online dataset $\mathcal{D}_{\mathrm{online}}$. This architecture enables data reuse under the same sparse, acceptance-aligned reward in an actor–learner design analogous to HIL-SERL [26,65].

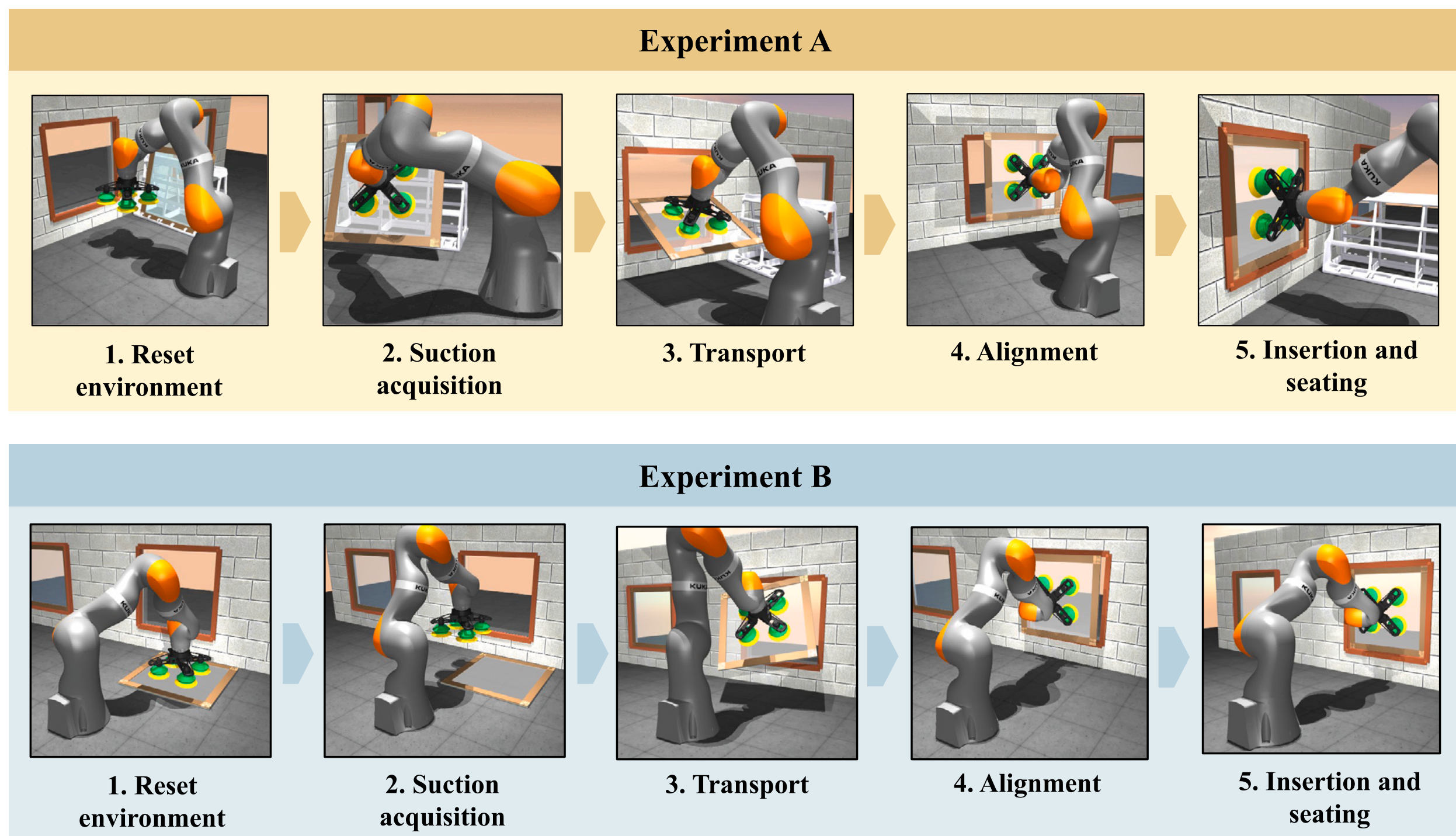

**Fig. 6.** Sequential milestones of the window installation workflow for both experimental settings. (a) Experiment A (precision insertion) isolates the tolerance-critical bottleneck by initiating from a structured staging pose. (b) Experiment B (end-to-end installation) evaluates the complete pipeline starting from randomized ground placement, introducing upstream uncertainty in suction acquisition and transport.

#### *3.4.1. Formalizing tacit knowledge via demonstration bootstrapping*

Prior to online fine-tuning, a skilled operator teleoperates the robot with a 6-DoF SpaceMouse to collect a fixed demonstration buffer $\mathcal{D}_{\text{demo}}$ of 50 trajectories spanning the complete workflow. The trajectories are recorded across varied initial configurations to cover approach, vacuum engagement, alignment, insertion, and seating under diverse contact conditions. They encode stage-dependent manipulation strategies that resist analytical specification, such as contact-aware micro-adjustments and recovery actions under partial observability, and are not directly transferable from nominal BIM geometry. Pretraining on $\mathcal{D}_{\text{demo}}$ initializes the policy within an installer-verified region of the action space and reduces early exploration into low-recoverability contact states such as deep wedging, thereby improving sample efficiency in offline-to-online RL [19,25,66,67]. This bootstrapping is especially consequential because the sparse terminal reward provides negligible learning signal until the policy reliably reaches late-stage contact regimes. All demonstrations were collected by the same operator who subsequently performed online interventions, ensuring consistency in recovery strategy and takeover timing across training runs.

#### *3.4.2. Active safety-aware online intervention*

During online training, the installer maintains an active safety–recoverability envelope through event-driven interventions. Following HIL-SERL [26], interventions are implemented as binary takeovers at the shared 10 Hz action interface with no action blending, and mid-chunk takeovers follow the transition-construction protocol defined in Section 3.2. Takeovers are initiated on the basis of observable cues at the shared interface, including loss of visible clearance, stalled insertion progress, asymmetric contact indicative of wedging, unstable suction engagement, repeated reinsertion attempts, or prolonged non-progressing contact states. These cues are assessed exclusively from the same RGB and proprioceptive observations available to the policy. Each takeover is timestamped to enable precise supervision burden accounting.

Within this protocol, two intervention types are distinguished, sharing the same overriding mechanism but differing in intent: (i) *safety takeovers*, which preempt or arrest imminent low-recoverability outcomes such as escalating contact, deep wedging, or constraint configurations unlikely to resolve within the episode horizon; and (ii) *coverage interventions*, applied sparingly to expose rare but practically consequential boundary regimes under controlled, low-risk conditions (e.g., mild misalignment near the opening) and elicit recovery trajectories that autonomous exploration is unlikely to encounter within a limited interaction budget. In both cases, interventions remain bounded by the safety–recoverability envelope and are not deployed to correct mild sub-optimality.

Although interventions are treated as standard off-policy data, they exert a systematic effect under sparse terminal rewards: they concentrate the replay distribution on high-consequence boundary states and recovery segments that dominate field failures, including unstable vacuum engagement, clearance loss during transport, repeated alignment failures, and contact-induced binding during insertion. Relative to unguided exploration, this yields higher-information interaction while reducing time spent in low-return, low-recoverability regimes [53]. Intervention frequency is typically high early in training and decreases as policy competence develops, producing progressive autonomy transfer under the same binary takeover protocol [26]. Intervention segments are logged exclusively to $\mathcal{D}_{\text{online}}$, while $\mathcal{D}_{\text{demo}}$ is held fixed [56,66]. Fig. 7 summarizes the training-time data flow with interventions and the evaluation-time protocol without interventions.

Offline-to-online fine-tuning for tolerance-critical assembly is susceptible to critic miscalibration under distribution shift, since demonstrations may not cover the discontinuous contact impulses and mode transitions encountered during autonomous execution. To mitigate this early instability, a three-stage protocol is employed that separates behavioral initialization, value calibration, and online fine-tuning with interventions.

**Stage I (behavioral prior initialization, offline pretraining):** The chunk-level actor and critic are pretrained on the static demonstration dataset $\mathcal{D}_{\text{demo}}$. This initializes the policy within a physically plausible, installer-verified behavior manifold and anchors value estimates to demonstrated contact dynamics, providing a conservative starting point for online interaction (Section 3.5).

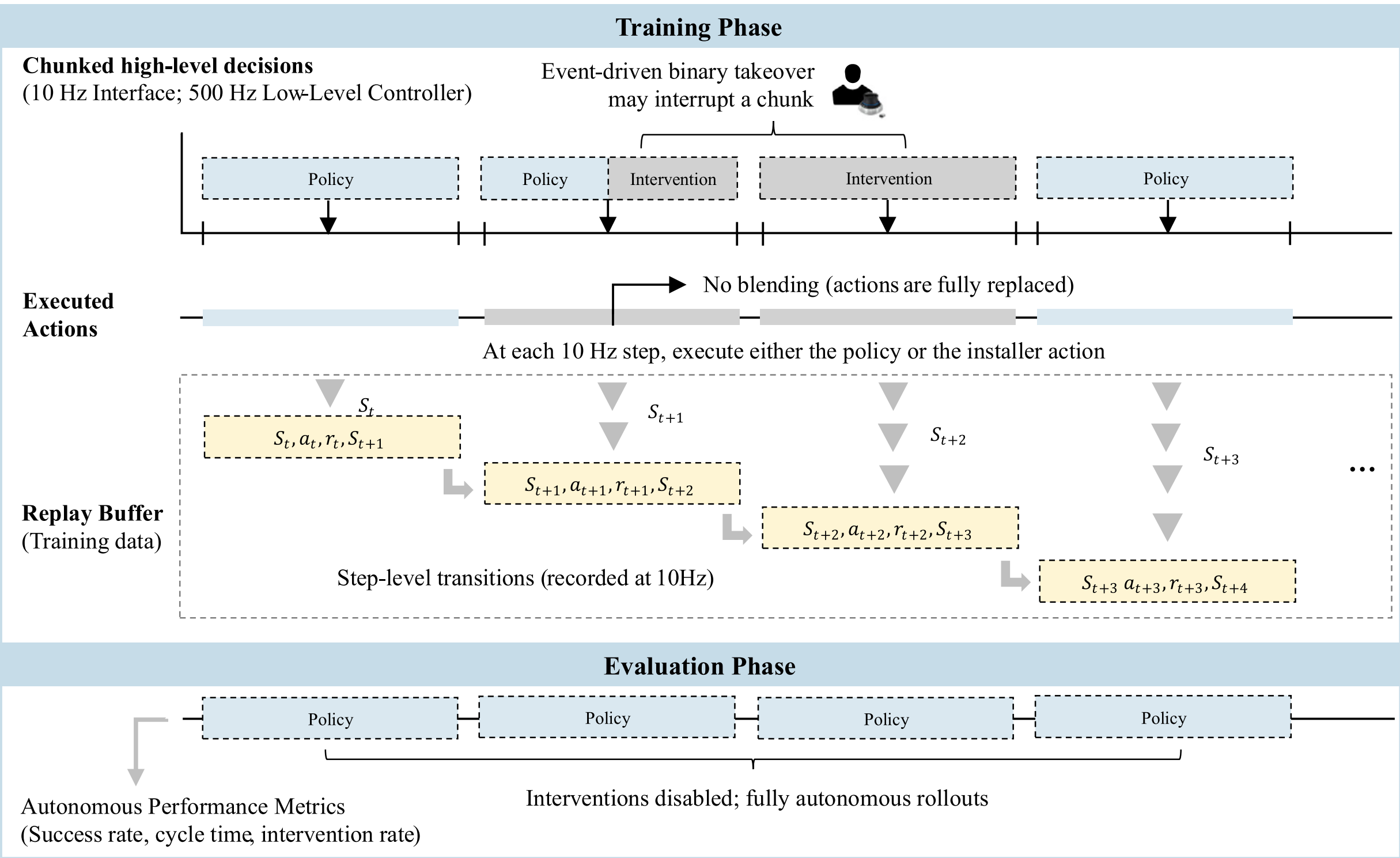

**Fig. 7.** Execution timeline: training phase (top) with chunked high-level decisions at 10 Hz (Policy vs. Intervention), executed actions (no blending), and step-level transitions $(s_t, a_t, r_t, s_{t+1})$ into the replay buffer ($a_t$ = actual executed action); evaluation phase (bottom) with interventions disabled and fully autonomous rollouts for success rate and cycle time.

**Stage II (warm-start data collection, non-updating value calibration):** Following behavioral initialization, online interaction commences while gradient updates are temporarily suspended for $N_{\text{warm}}$ transitions. Interaction data are stored in $\mathcal{D}_{\text{online}}$ without parameter updates. This phase populates $\mathcal{D}_{\text{online}}$ with on-policy interaction experience, so that once updates begin, early critic learning is less dominated by the demonstration-only distribution. The result is a reduction in early value miscalibration and a corresponding decrease in unnecessary intervention during the offline-to-online transition [12,56,57].

**Stage III (online fine-tuning with interventions):** Gradient updates are enabled and training proceeds with concurrent data collection. The online replay buffer $\mathcal{D}_{\text{online}}$ accumulates autonomous rollouts and event-driven intervention segments, and training batches are drawn from the mixed dataset $\mathcal{D} = \mathcal{D}_{\text{demo}} \cup \mathcal{D}_{\text{online}}$. Interventions are treated as standard off-policy transitions, including mid-chunk overrides, so learning optimizes value estimates over executed action sequences while preserving temporal abstraction. Following the recommendations of RLPD [56], a $1:1$ sampling ratio is maintained between offline expert demonstrations and online interaction experience. This strategy keeps the policy anchored to expert behaviors while optimizing for online task success, mitigating the catastrophic forgetting characteristic of naive offline-to-online transitions.

### *3.5. Generative action-sequence policy for long-horizon assembly with sparse rewards*

The learner optimizes the policy within the chunked action space $\mathcal{A}^h$ defined in Section 3.2. To represent the inherent multimodality of recovery maneuvers, QC-FQL is adopted via a teacher–student parameterization [12]: a generative flow-based prior $f_\xi$ (teacher) characterizes the action-sequence manifold, while a one-step noise-conditioned actor $\mu_\psi$ (student) enables real-time, constant-time control. Compared to diffusion-based policies [68], QC-FQL factorizes multimodal behavior modeling and value optimization through teacher–student distillation, eliminating iterative sampling during online action selection. This decoupling is particularly advantageous under sparse terminal rewards, where critic instability can otherwise corrupt generative sampling and destabilize recovery behavior. A common alternative is step-wise RL with $n$-step targets; in offline-to-online settings, however, such targets can incur elevated off-policy bias when the buffered action suffix deviates from the current policy. QC-FQL instead defines value on the executed action chunk $Q(s_{kh}, \mathbf{u}_k)$ and applies a one-step TD backup in the induced SMDP, preserving target–input alignment. At each decision step $k$ (environment time $t = kh$), the policy outputs a length-$h$ action sequence $\mathbf{u}_k \in \mathcal{A}^h$ from state $s_{kh}$. The sequence is executed for $h$ steps to produce the accumulated discounted chunk return $R_k^{(h)}$ and effective discount $\Gamma = \gamma^h$. As illustrated in Fig. 8, the critic $Q_\phi(s_{kh}, \mathbf{u}_k)$ is updated on this executed sequence, avoiding the off-policy mismatch between the value input and the return it generated.

Formally, this yields a one-step Bellman backup with effective discount $\Gamma$:

$$Q(s_{kh}, \mathbf{u}_k) = R_k^{(h)} + \Gamma\, \mathbb{E}_{\mathbf{u}' \sim \mu(\cdot|s_{kh+h})} \left[ Q(s_{kh+h}, \mathbf{u}') \right], \tag{5}$$

where $\Gamma$ is the effective discount factor defined in Eq. (3) and the bootstrap term is masked upon episode termination within the chunk. The TD target $y_k$ and corresponding critic loss $\mathcal{L}_Q(\phi)$ are

$$y_k = R_k^{(h)} + \Gamma\, \mathbb{E}_{z \sim \mathcal{N}(0,I)} \left[ Q_{\bar{\phi}}(s_{kh+h}, \mu_\psi(s_{kh+h}, z)) \right], \tag{6}$$

$$\mathcal{L}_Q(\phi) = \mathbb{E}\left[ \left( Q_\phi(s_{kh}, \mathbf{u}_k) - y_k \right)^2 \right], \tag{7}$$

where $\bar{\phi}$ denotes the target network parameters. In the implementation, the student actor $\mu_\psi$ is used for bootstrapping to ensure consistency between the learned value and the executable policy.

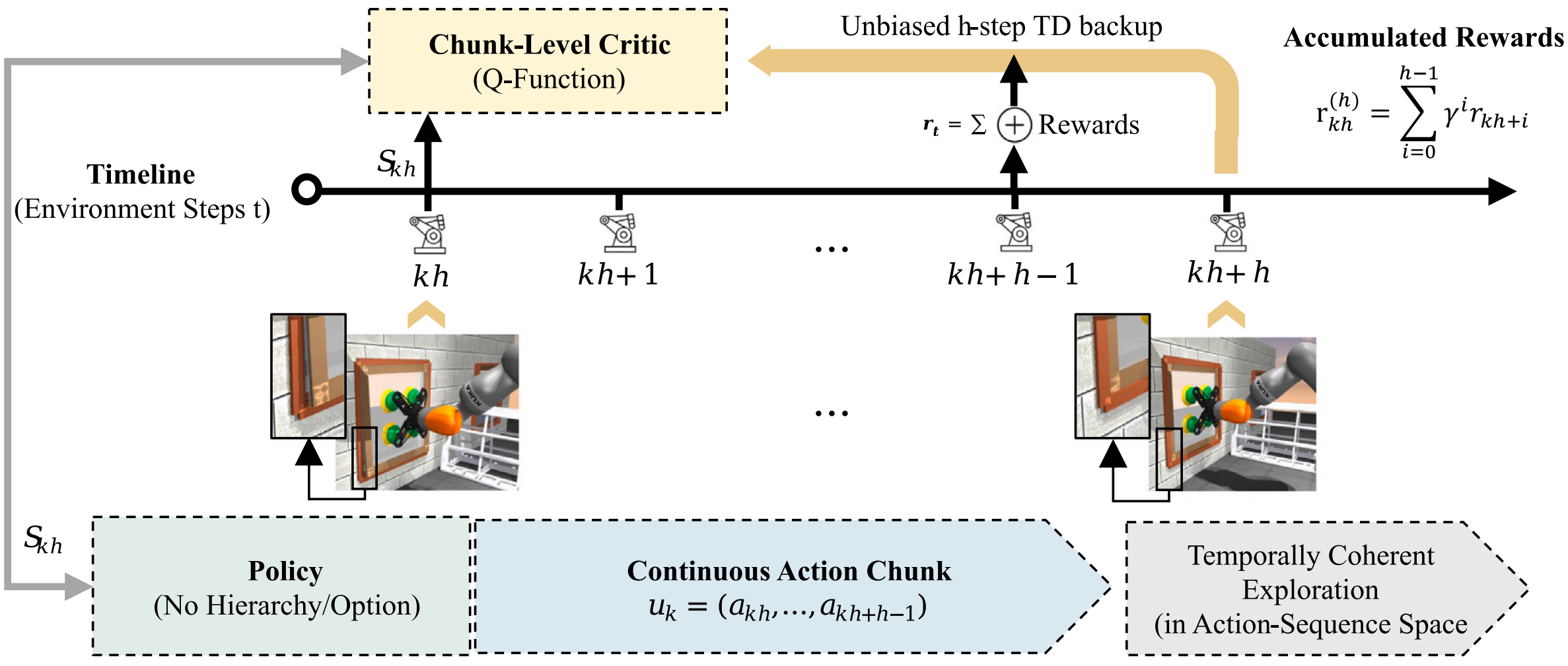

**Fig. 8.** Chunked Bellman backup in the induced SMDP. At decision step $k$, the critic $Q_\phi(s_{kh}, \mathbf{u}_k)$ is updated on the executed length-$h$ action sequence $\mathbf{u}_k$ using the chunk return $R_k^{(h)}$, pre-computed during transition construction in the replay buffer as the accumulated discounted sum over the $h$-step sequence, and the effective discount $\Gamma = \gamma^h$.

*Multimodal sequence prior via a behavior cloning flow policy.* The chunked action space $\mathcal{A}^h$ accommodates multimodal recovery under frictional contact, because distinct contact modes often require mutually exclusive corrective sequences. In tight-tolerance assembly, such maneuvers are intrinsically multimodal: for instance, pivoting left versus right to resolve a bind. Single-Gaussian policies commonly employed in step-wise RL (e.g., SAC, PPO) tend to average across conflicting corrective modes, producing smooth but mechanically ineffective corrections that can exacerbate jamming under sparse rewards. The flow-based behavior prior $f_\xi$ preserves the multimodality of the action-sequence space, enabling the student $\mu_\psi$ to represent and execute distinct, physically plausible recovery trajectories required for navigating narrow clearances [12,69].

To represent these modes while preserving real-time control capability, the QC-FQL teacher–student architecture is adopted. The behavior cloning flow policy $f_\xi$ (teacher) is trained via a Rectified Flow objective [70] on replay data to approximate the multimodal behavior distribution over the action-sequence space. Specifically, the training objective minimizes the squared error between the predicted velocity field and the constant velocity connecting a noise sample $x_0 \sim \mathcal{N}(0, I)$ to a data sample $x_1$ (the action chunk) at randomly sampled interpolation times $t \sim \mathcal{U}(0, 1)$:

$$\mathcal{L}_{\mathrm{RF}}(\xi) = \mathbb{E}_{x_0 \sim \mathcal{N}(0,I), x_1 \sim \mathcal{D}, t \sim \mathcal{U}(0,1)} \left[ \left\| f_\xi(x_t, t, s) - (x_1 - x_0) \right\|^2 \right], \tag{8}$$

where $x_t = (1-t) \cdot x_0 + t \cdot x_1$ is the linear interpolation path. A behavior chunk sample is obtained by integrating the learned velocity field from a Gaussian noise seed $z \sim \mathcal{N}(0, I)$ using an ODE solver (e.g., the Euler method with $K$ steps), denoted $\mathbf{a}^\xi(s, z)$. To avoid the computational overhead of iterative sampling during online RL, this prior is distilled into a one-step noise-conditioned actor $\mu_\psi(s, z)$ (student).

This factorization reduces execution-time sampling cost by eliminating iterative integration and decouples multimodal behavior modeling from value-driven policy optimization. The separation directly addresses a failure mode in tolerance-critical installation: the action-sequence space is non-convex and multimodal, and interpolation between incompatible recovery maneuvers can intensify binding under frictional contact. The flow-based behavior model $f_\xi$ thus serves as a multimodal sequence prior that regularizes the student toward replay-supported recovery manifolds, while chunk-level Q-learning selects among modes according to expected return. This teacher–student factorization follows the QC-FQL formulation [12], wherein the behavior model supplies the multimodal prior and the one-step actor enables constant-time control.

*Chunk-level policy optimization with behavior-model distillation.* The student actor $\mu_\psi$ is optimized to maximize expected chunk-level value while remaining proximate to the behavior model $f_\xi$ via a teacher–student regularizer. Following the QC-FQL objective [12,69], the student minimizes:

$$\mathcal{L}_\pi(\psi) = \mathbb{E}_{s \sim \mathcal{D},\, z \sim \mathcal{N}(0,I)} \left[ -Q_\phi\big(s, \mu_\psi(s,z)\big) + \alpha \left\| \mu_\psi(s,z) - \mathbf{a}^\xi(s,z) \right\|_2^2 \right], \tag{9}$$

where $\mathbf{a}^\xi(s,z) \in \mathcal{A}^h$ denotes the target action chunk generated by the flow-based behavior prior $f_\xi$. The squared $\ell_2$ distance is computed over the concatenated chunk dimension ($h \times d_{\mathrm{action}}$) rather than as a step-wise average. This formulation acts as a distribution-level regularizer that prevents the student policy $\mu_\psi$ from drifting away from the replay-supported behavior manifold. By penalizing deviations across the full temporal horizon $h$, the loss encourages coherent multi-step recovery manifolds, such as coordinated retract-and-tilt sequences, rather than isolated step-wise action matching. During optimization, $f_\xi$ provides detached distillation targets, allowing the generative prior to preserve the multimodality of installer expertise while the distilled student $\mu_\psi$ remains a reactive policy suited to real-time inference during high-precision assembly.

### 3.6. System implementation and training protocol

The computational architecture couples the execution pipeline with the QC-FQL learning core. The system is implemented in PyTorch and integrated with MuJoCo, with hyperparameters provided in Appendix. The control policy processes dual-view RGB inputs alongside robot proprioception to output $h$-step end-effector pose increments.

*Visual representation and sensor fusion.* Two simulated RGB cameras provide a wrist-mounted view for fine alignment and a side view for global context. The visual backbone is a ResNet-10 encoder pre-trained on ImageNet [71]. Visual features are fused with the proprioceptive state $x_t$ (joint kinematics $q$, $\dot{q}$ and end-effector pose $p_{\mathrm{ee}}$) via an MLP. To mitigate representation drift during the offline-to-online transition, the ImageNet-pretrained ResNet-10 visual encoder is kept frozen throughout training. Actor and critic updates are applied only to the downstream fusion, policy, teacher, and value networks.

*Hierarchical control and policy execution.* The algorithmic core implements a teacher–student architecture in which a flow-matching teacher $f_\xi$ provides distillation targets for a one-step, noise-conditioned student actor $\mu_\psi$. At each decision index $k$, the student outputs a flattened vector of dimension $h \times d_{\text{action}}$ ($d_{\text{action}} = 7$). This vector is reshaped into a temporal chunk and executed sequentially at 10 Hz. A 500 Hz OSC serves as a tracking layer bridging high-level chunked decisions and low-level robot dynamics. Rather than employing explicit force control, effective compliance emerges from the interaction between the tracking stiffness of the OSC and the contact-rich environment.

*Training protocol and stability.* Learning is stabilized via an ensemble of $N = 2$ critics with mean-aggregation for TD targets, mitigating overestimation bias during online fine-tuning. Training follows the three-stage protocol detailed in Section 3.4, with a warm-start buffer of $N_{\text{warm}} = 10{,}000$ transitions and an Update-to-Data (UTD) ratio of 2. Truncated transitions (e.g., at the episode horizon) are excluded from the critic TD loss to prevent bootstrap targets from being applied beyond the task horizon.

## 4. Experimental results

The evaluation defines the experimental setup and protocol, then presents quantitative results, ablations, and end-to-end performance analyses.

### 4.1. Experimental setup

The proposed installer-in-the-loop framework is evaluated in the MuJoCo benchmark introduced in Section 3.3 for clearance-limited, contact-rich window installation. The experimental design addresses three questions of direct relevance to construction automation: (i) whether the system can reliably acquire the complete workflow, including terminal clearance-limited insertion and seating, under sparse acceptance-aligned rewards; (ii) how much online interaction time and human supervision are required to reach stable autonomy; and (iii) which system components are necessary for sample-efficient, stable learning.

Two evaluation settings are considered (Fig. 6). They share identical insertion–seating mechanics but differ in upstream uncertainty and overall horizon length. Experiment A assumes a deployment-relevant staging workflow in which the window unit is pre-positioned in a repeatable pose (e.g., on a staging rack) with only small perturbations introduced. This setting isolates the clearance-limited insertion–seating bottleneck under controlled but realistic variability. Experiment B removes the staging assumption by initializing directly from randomized ground placement, thereby introducing additional uncertainty in suction acquisition and coarse alignment before the same contact-rich insertion phase. This setting evaluates workflow-level integration by examining how upstream variability propagates into seating outcomes and how the framework sustains performance under cumulative geometric uncertainty across configurations.

*Experiment a: Precision insertion under structured staging.* Experiment A evaluates sample efficiency and safety under a controlled staging condition in which the window panel is placed in a repeatable pose representative of on-site preparation (e.g., a tilted staging rack), consistent with common safety guidance for storing window panels on inclined racks [72]. Small perturbations are applied around the staged configuration to reflect realistic placement noise while controlling upstream acquisition variability, thereby isolating the clearance-limited insertion–seating bottleneck.

*Experiment b: End-to-end installation from ground placement with randomized planar pose.* Experiment B removes the structured staging assumption and evaluates end-to-end installation from ground placement with fully randomized initial poses. The initial window pose is randomized in the horizontal plane: position is sampled within a $0.2\,\text{m} \times 0.6\,\text{m}$ rectangular horizontal region, and orientation is perturbed by a yaw rotation sampled from a symmetric range of $\pm 15^{\circ}$ about the vertical axis. The robot must acquire suction before insertion and seating, introducing additional perception uncertainty and grasp variability. Stage-wise outcomes are reported in Table 4, decomposing each episode into suction acquired, insertion entered, and successfully seated, to avoid confounding suction acquisition failures with insertion/seating failures.

*Scope of claims.* All empirical performance claims are explicitly conditioned on the variability ranges and contact-modeling assumptions specified in Section 3.3. In particular, robustness results are evaluated under bounded pose perturbations, friction randomization, and compliant contact parameters within stable simulation regimes. No claim of robustness beyond these ranges is made (e.g., extreme friction leading to deep self-locking, large initial misalignment outside the randomized bounds, or unmodeled structural compliance). Reported results should therefore be interpreted as performance within a defined stress-test regime rather than as guarantees under arbitrary construction variability.

### 4.2. Evaluation protocol and metrics

Across both experiments, progress is reported against wall-clock online interaction time, which includes simulator stepping, rendering, policy inference, data logging, and any overhead attributable to installer intervention. Although evaluation is conducted entirely in simulation, the installer participates in real time during training; wall-clock time therefore serves as a proxy for calendar-time adaptation and supervision burden in deployment-oriented simulated workflows. All methods involving environment interaction are trained under a fixed 3.0 h wall-clock online interaction budget on a single workstation (Ubuntu 20.04, Intel Core i9-14900F, NVIDIA RTX 4070 SUPER). All variants using prior data are initialized with the same set of 50 teleoperated demonstration trajectories collected under an identical sensing, control, and termination interface. Offline-only baselines perform no online interaction and are evaluated immediately after offline training.

*Training budget, checkpoints, and stochasticity.* The 3.0 h budget is enforced in wall-clock real time at the environment step rate (10 Hz), so all online methods consume an identical number of environment steps. Policies are evaluated on a fixed wall-clock checkpoint schedule aligned with the training logs; learning curves in figures employ running averages over training episodes as described below. Training uses fixed simulator and algorithm random seeds for reproducibility. Tabulated results report one standardized training session per method variant under this protocol. During pilot development, each online variant was run at least twice under identical hyperparameters to verify qualitative convergence behavior; these pilot runs are not reported as independent training seeds because operator participation cannot be held fixed across sessions.

**Wall-clock accounting, step budget, and checkpoints.** Unless episodes terminate early, advancing the simulator at 10 Hz for the full 3.0 h wall-clock online interaction window (including warm-start when enabled) corresponds to up to $3.0 \times 3600 \times 10 \approx 1.08 \times 10^{5}$ high-level environment transitions per online method. Early terminations alter the number of completed episodes but not the fixed real-time stepping budget. Episode counts therefore vary with task duration and resets (Appendix: max horizon 600 steps). Policies are evaluated at checkpoints recorded on a fixed wall-clock grid aligned with the training logs; successive evaluation checkpoints are spaced by approximately five minutes of wall time during online training (subject to logging overhead). Human effort is aggregated from 10 Hz takeover steps and constitutes a small fraction of total wall time in the reported runs because interventions are event-driven rather than continuous; the dominant computational costs are simulator stepping, rendering, and learning-side forward/backward passes.

*Autonomous evaluation and curve smoothing.* At each evaluation checkpoint, the policy is frozen and assessed over $M = 100$ rollouts under varied initial conditions (small perturbations in Experiment A; randomized ground placement in Experiment B), with installer interventions and manual safety terminations disabled. Each checkpoint yields one success-rate estimate, computed as the mean over these $M = 100$ fully autonomous rollouts. All success and cycle-time statistics reported in tables are derived exclusively from autonomous evaluation rollouts. For visualization, learning curves display success-rate and cycle-time statistics computed from training episodes, which may include installer interventions, and smoothed with a running average over the most recent 20 training episodes, following HIL-SERL [26]. This approach provides a continuous view of learning progress during training, while evaluation checkpoints deliver unbiased estimates of autonomous policy performance. The evaluation sample size $M = 100$ was selected to achieve a margin of error of $\pm 9.8$ percentage points at the 95% confidence level for binary success outcomes (conservative normal approximation: $n = z^2 \cdot 0.25/e^2$, $z = 1.96$); at the observed success rates of 92% and 100%, the corresponding 95% Wilson score intervals are $[85.0\%, 95.9\%]$ and $[96.3\%, 100\%]$, indicating that reported performance gaps exceed sampling variability.

*Milestone metric.* Time-to-95% is reported as an early-performance milestone, defined as the earliest wall-clock time (measured from the onset of online interaction, including warm-start when enabled) at which the raw autonomous success rate reaches at least 95% for $J = 3$ consecutive evaluation checkpoints. All methods are trained for the full 3.0 h budget; this metric does not imply early stopping.

*Cycle time.* Cycle time is the mean duration from episode start to successful seating, computed over successful autonomous evaluation rollouts; methods with 0% success yield no cycle-time value.

*Intervention-related metrics.* Intervention statistics are computed during training under the binary takeover protocol, wherein teleoperation fully replaces the policy action until control is returned. For intervention-rate curves, a sliding-window estimate over the most recent 20 training episodes is used, reporting the fraction of 10 Hz action steps executed under human takeover. Human effort (in minutes) is derived by converting the number of 10 Hz intervention steps to wall-clock duration and aggregating over the full training run.

*Implementation fairness and computational parity.* All compared methods share identical observation modalities (dual-view RGB and proprioception), the same frozen visual encoder, the same low-level operational-space tracking controller, identical termination criteria, and the same demonstration dataset. Online methods operate under the same 3.0 h wall-clock interaction budget on identical hardware, and wall-clock time encompasses simulator stepping, rendering, policy inference, and intervention overhead. No method is permitted to exceed the real-time stepping rate of the environment. Hyperparameters for each baseline follow either the original publication recommendations (e.g., HIL-SERL) or standard implementations (e.g., SAC), without task-specific tuning beyond what is disclosed. Consequently, all online variants accumulate an identical number of environment interaction steps, independent of policy inference latency. This controlled protocol supports attributing observed performance differences to algorithmic design rather than computational advantage or sensing discrepancies.

### 4.3. Baselines and ablations

Three external baselines and a set of controlled ablations are included to attribute observed gains to installer intervention, warm-start value calibration, and temporal abstraction. Unless noted otherwise, all variants share identical observation modalities, the low-level controller, termination rules, the demonstration dataset, and the fixed 3.0 h wall-clock online interaction budget. Action Chunking with Transformers (ACT) [73] is an imitation-learning reference trained exclusively on $\mathcal{D}_{\text{demo}}$ with no online interaction or RL updates. Standard SAC [41] employs step-wise actions ($h = 1$), initializes the replay buffer with $\mathcal{D}_{\text{demo}}$, and performs neither offline pretraining nor human intervention during online learning. Following Luo et al. [26], HIL-SERL uses the SAC/RLPD pipeline with offline pretraining, non-updating warm-start, and binary takeovers, with step-wise control ($h = 1$) and a unimodal Gaussian policy, isolating human-guided exploration from the effect of chunking. An additional step-wise variant with matched installer interventions, *Ours (step-wise + intervention)*, is included to isolate the combined effect of chunking and intervention. The ablated variants include: step-wise QC-FQL ($h = 1$) with offline initialization and warm-start but no intervention; a flow-based behavior prior trained only on $\mathcal{D}_{\text{demo}}$ with no online updates; offline pretraining followed by warm-start rollout collection with parameters frozen for the full budget; the full pipeline without warm-start; the full chunked pipeline ($h = 10$) with warm-start but without binary takeovers during online fine-tuning; and the complete installer-in-the-loop framework with offline initialization, warm-start, $h = 10$, and sparse interventions.

To attribute performance gains to individual system components, controlled ablations vary exactly one factor at a time while holding the demonstration dataset, observation modalities, low-level controller, termination rules, and fixed 3.0 h budget constant. Temporal abstraction is isolated by comparing the chunked variant with $h = 10$ actions (*Ours w/o intervention*) against its step-wise counterpart with $h = 1$, revealing the impact of temporally extended action sequences on long-horizon learning under sparse terminal rewards. Installer-guided exploration is assessed by comparing the complete pipeline against a no-intervention variant with warm-start calibration and temporal abstraction held fixed; this isolates the effect of binary control-sharing takeovers on safety-aware data collection and recovery learning. Warm-start value calibration is evaluated by contrasting the complete pipeline with a variant that transitions directly from offline initialization to online fine-tuning, while keeping interventions and temporal abstraction unchanged. This isolates the role of non-updating warm-start interaction in stabilizing offline-to-online transfer under contact-rich dynamics. Table 1 summarizes which components are active in each method.

### 4.4. Main results of Experiment A

Fig. 10 and Table 2 summarize Experiment A under structured staging, isolating clearance-limited insertion–seating while controlling upstream variability. All online methods operate under the same 3.0 h wall-clock budget; time-to-95% is treated as an early adaptation milestone, while all runs continue to the full budget to assess post-threshold stability. Under this matched protocol and without evaluation-time intervention, the complete pipeline provides the strongest evidence of reliable autonomous seating: 100% final seated success, milestone crossing at ~0.5 h, and a 27.1 s cycle time. HIL-SERL converges to 92%, while SAC remains at 0%, consistent with the well-documented difficulty of sparse-reward learning in long-horizon contact dynamics. Temporal abstraction is decisive in the no-intervention condition: step-wise control ($h = 1$) yields 0% success versus 87% for chunked control ($h = 10$), although the chunked no-intervention variant remains less stable than the complete pipeline. With matched installer interventions enabled, the step-wise QC-FQL variant reaches 53% final success (cycle time 32.7 s), substantially above the no-intervention step-wise baseline yet still below chunked variants. The supported interpretation is therefore component-specific: intervention improves recoverability, but does not substitute for temporal abstraction in this clearance-limited regime. Fig. 9 illustrates a representative retreat–pivot–reinsert recovery from a jamming event.

Relative to the no-intervention variant, the complete pipeline exhibits fewer prolonged success-rate drops and more stable post-threshold performance, consistent with takeovers preventing early excursions into low-recoverability contact regimes during adaptation.

**Table 1**
Component activation matrix for external baselines and controlled ablations. Columns indicate whether each protocol component is enabled; ✓denotes use of the component and ×denotes its removal.

| Method | Demo | Online RL Updates | Intervention | Warm-start | Chunking |
|---|---|---|---|---|---|
| ACT (offline IL) | ✓ | × | × | × | ✓ |
| SAC (online RL) | ✓ | ✓ | × | × | × |
| HIL-SERL | ✓ | ✓ | ✓ | ✓ | × |
| Ours (step-wise) | ✓ | ✓ | × | ✓ | × |
| Ours (step-wise + intervention) | ✓ | ✓ | ✓ | ✓ | × |
| Ours (offline-only) | ✓ | × | × | × | ✓ |
| Ours (offline + warm-start) | ✓ | × | × | ✓ | ✓ |
| Ours (w/o warm-start) | ✓ | ✓ | ✓ | × | ✓ |
| Ours (w/o intervention) | ✓ | ✓ | × | ✓ | ✓ |
| Ours (complete pipeline) | ✓ | ✓ | ✓ | ✓ | ✓ |

**Table 2**
Experiment A sample-efficiency and supervision burden under the structured-staging protocol. Success Rate is measured over $M = 100$ fully autonomous evaluation rollouts per checkpoint with interventions disabled. Online Interaction Time denotes the fixed 3.0 h wall-clock budget, including warm-start when enabled. Time-to-95% is the earliest point at which raw success exceeds 95% for $J = 3$ consecutive checkpoints. Human Effort is cumulative installer takeover duration during training, not evaluation-time assistance.

| Method | Final Success | Online Interaction Time | Time-to-95% | Cycle Time | Human Effort |
|---|---|---|---|---|---|
| | (%) | (h) | (h) | (s) | (min) |
| ACT (offline IL) | 62 | – | – | 31.9 | 0 |
| SAC (online RL) | 0 | ~3.00 | – | – | 0 |
| HIL-SERL | 92 | ~3.00 | – | 29.6 | 20 |
| Ours (step-wise) | 0 | ~3.00 | – | – | 0 |
| Ours (step-wise + intervention) | 53 | ~3.00 | – | 32.7 | 17 |
| Ours (offline-only) | 78 | – | – | 30.4 | 0 |
| Ours (offline + warm-start) | 85 | ~3.00 | – | 30.3 | 0 |
| Ours (w/o warm-start) | 90 | ~3.00 | – | 29.4 | 40 |
| Ours (w/o intervention) | 87 | ~3.00 | – | 28.1 | 0 |
| **Ours (complete pipeline)** | **100** | ~3.00 | ~0.50 | **27.1** | **12** |

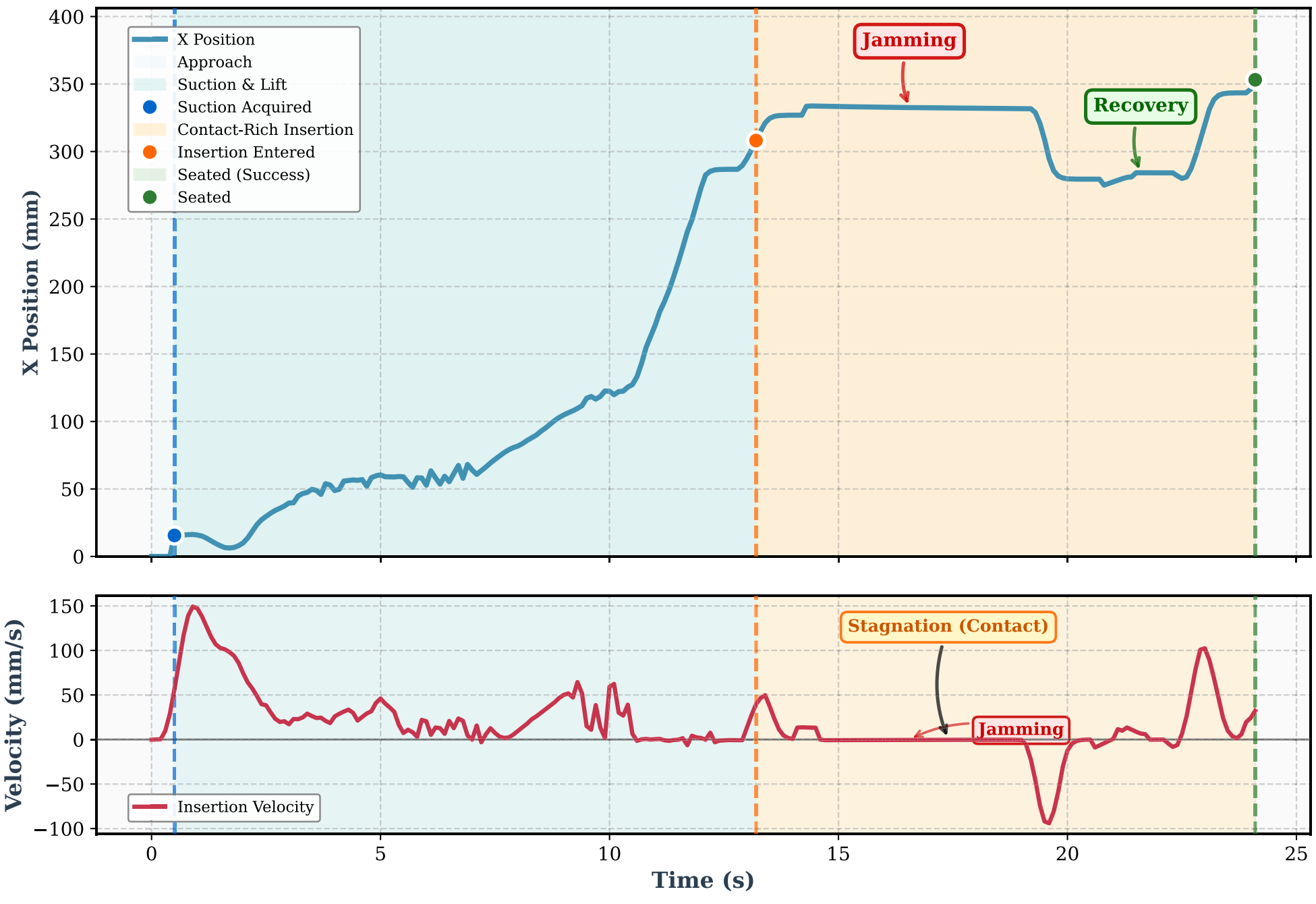

**Fig. 9.** Representative autonomous recovery behavior during clearance-limited insertion. Key events: *Suction Acquired* (1.5 s), *Insertion Entered* (13.5 s), *Jamming* (14.5 s) where positional progress stagnates and velocity drops to near zero, *Recovery* (19 s) involving partial retraction and reinsertion, and *Seated* success (22.5 s). The trajectory illustrates how the chunked action policy executes temporally coherent multi-step recovery sequences under sparse terminal rewards, resolving contact deadlocks through a coordinated retreat-and-reinsert maneuver.

Offline-only baselines remain insufficient (ACT 62%, *Ours (offline-only)* 78%), underscoring a persistent gap between demonstration-only behavior and reliable seating. In Experiment A, the intervention rate decays from ~0.25 to near zero by 0.5 h while autonomous success remains ≥95%, indicating that recovery is genuinely learned rather than continuously teleoperated. The complete pipeline also achieves lower supervision burden and reduced cycle time relative to HIL-SERL (12 vs 20 min takeover; 27.1 vs 29.6 s cycle time; Fig. 10 and Table 2). Consistent with Table 2, cycle-time improvement is not attributable to intervention alone: *Ours (w/o intervention)* already achieves 28.1 s versus 29.6 s for HIL-SERL under matched budget, indicating that temporal abstraction itself contributes to execution efficiency.

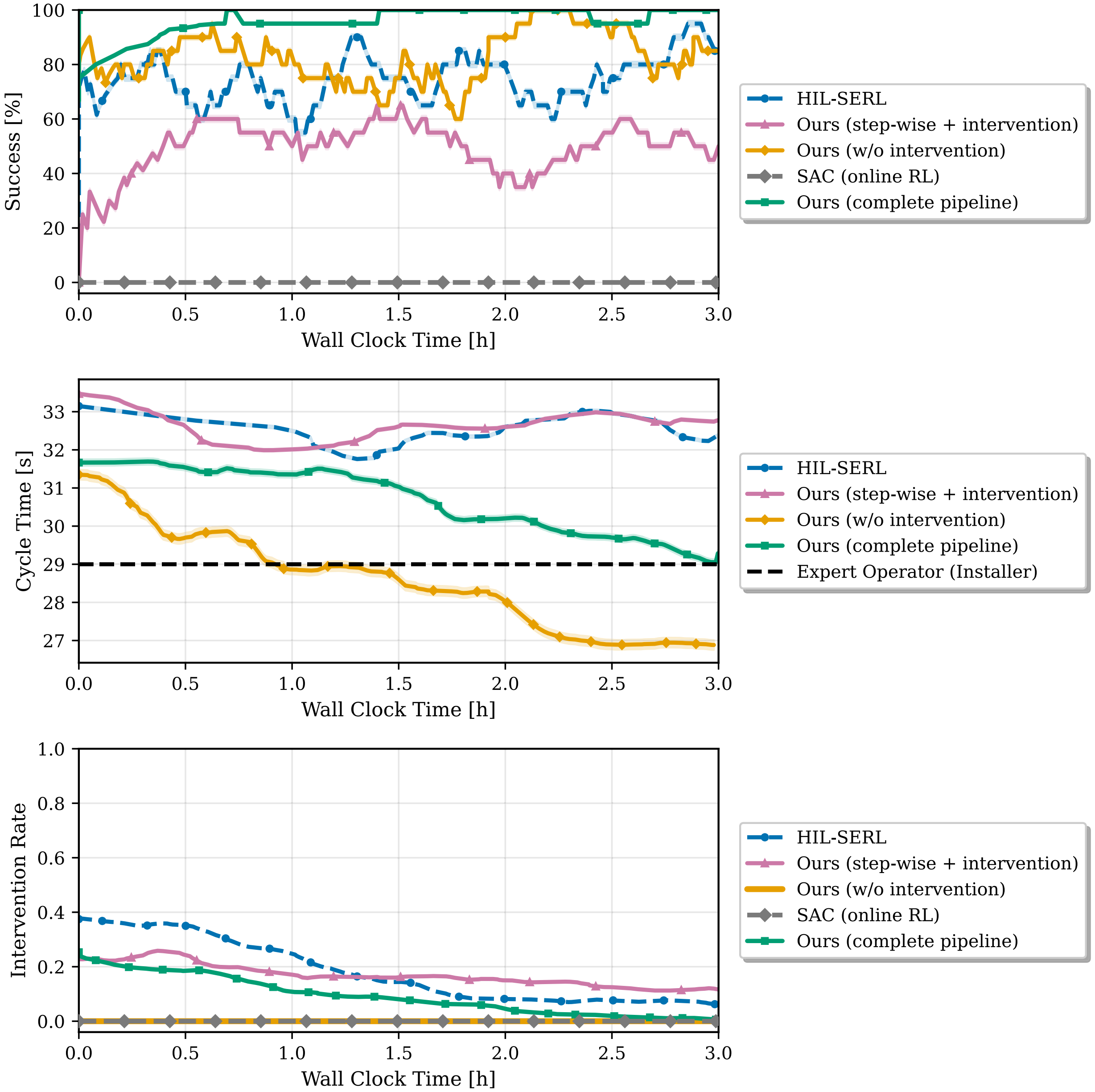

**Fig. 10.** Learning curves for Experiment A during online training. Curves are displayed as running averages over 20 episodes.

### 4.5. Experiment B: End-to-end installation under upstream uncertainty

Fig. 11 and Table 3–Table 4 summarize Experiment B, the end-to-end setting initiated from randomized ground placement (Section 4.1). Relative to Experiment A, this setting preserves the insertion–seating mechanics while extending the horizon and substantially increasing upstream uncertainty in suction acquisition and coarse alignment. All online methods operate under the same 3.0 h wall-clock budget; time-to-95% is reported as an early adaptation milestone, and all runs continue to the full budget for matched late-training comparison.

Under the same 3.0 h budget, the complete pipeline again provides the strongest evidence across reliability, adaptation speed, and supervision burden: 100% final seated success, milestone crossing at ∼1.5 h, and stable late-training performance (Fig. 11 and Table 3). HIL-SERL converges to 82%, while SAC and no-intervention step-wise control remain at 0%. With matched installer interventions enabled, the QC-FQL step-wise variant reaches 56% final success with stage-wise decomposition $\Pr(S) = 77.0\%$, $\Pr(I \mid S) = 79.2\%$, and $\Pr(T \mid I) = 90.3\%$, indicating that interventions recover substantial performance from the step-wise baseline but still underperform chunked variants under the same budget. Offline-only policies also degrade substantially (ACT 36%, ours offline-only 75%), reflecting distribution shift induced by upstream uncertainty. The ablations localize complementary contributions rather than merely ranking methods: removing interventions reduces post-insertion recoverability ($\Pr(T \mid I)$, Table 4), while removing warm-start increases the takeover burden (28 vs 15 min). The complete pipeline further achieves the lowest cycle time (18.8 s vs 21.9 s for HIL-SERL). Notably, as shown in the cycle-time curve of Fig. 11, the complete pipeline converges to cycle times below the expert operator baseline (dashed line) *in simulation*. This gap primarily reflects the elimination of human perception–decision latency in the teleoperation interface and the use of idealized sensing, and does not constitute evidence of superiority over skilled installers in physical field settings.

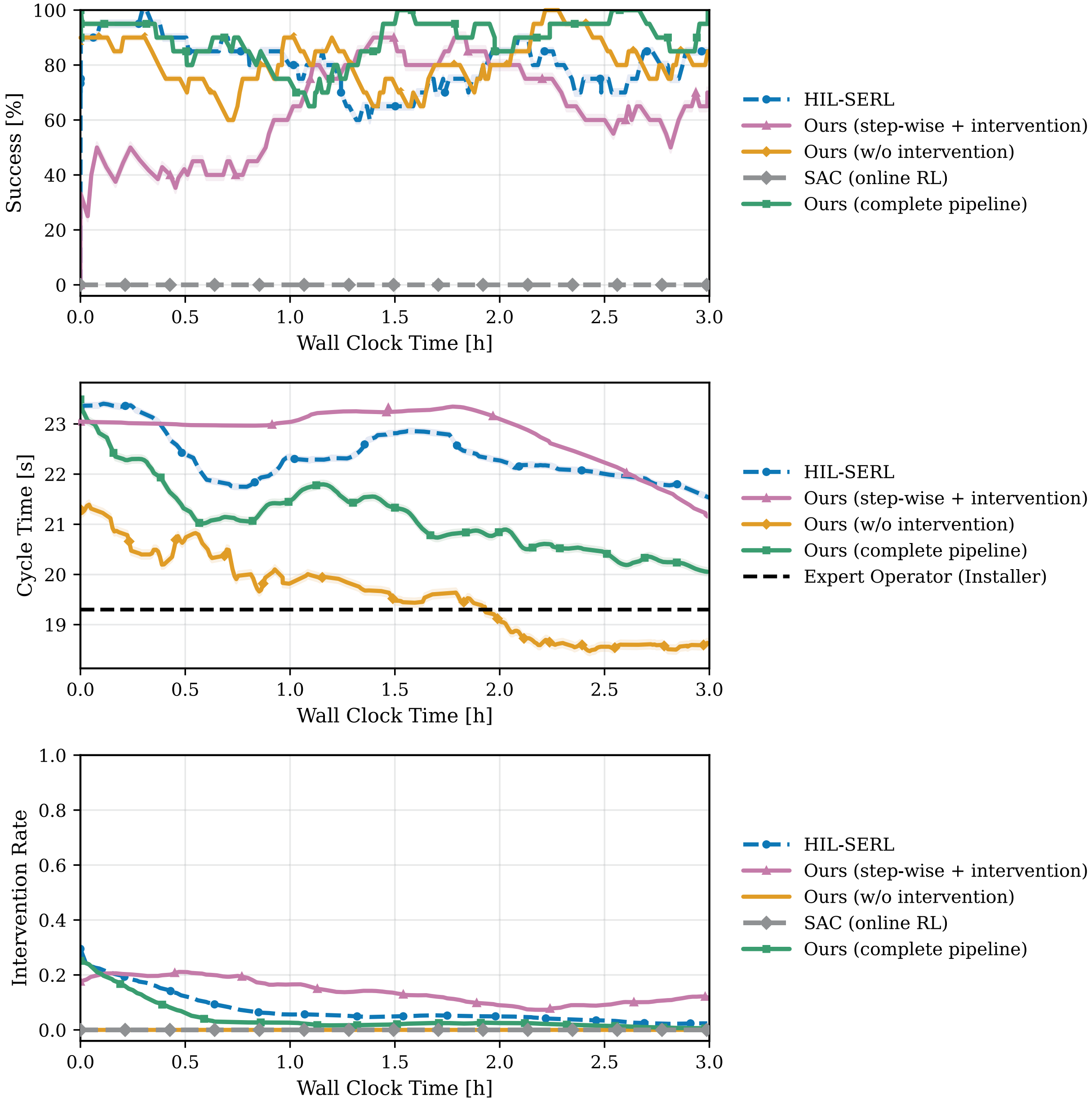

**Fig. 11.** Learning curves for Experiment B during online training. Curves are displayed as running averages over 20 episodes.

This behavior is consistent with Fig. 9, in which the policy resolves mid-depth jamming through a non-monotonic retraction–reinsertion maneuver that is inaccessible to unimodal step-wise control. The shorter Experiment B cycle time (18.8 s) relative to Experiment A (27.1 s) primarily reflects the difference in starting configuration: Experiment A initiates from a tilted staging rack and requires larger pre-insertion reorientation, whereas Experiment B initiates near-base from ground placement. The intervention-rate curve in Fig. 11 also decays toward zero, consistent with progressive autonomy transfer under the same takeover protocol.

Table 4 summarizes how upstream uncertainty reshapes the distribution of failure modes. Relative to Experiment A, a larger fraction of failures occurs before insertion: methods without robust acquisition behavior exhibit reduced suction success $\Pr(S)$, directly reducing the fraction of episodes that reach the clearance-limited insertion regime. Conditioned on suction, insertion entry is near-saturated for competitive methods ($\Pr(I \mid S) \approx$ 96.4–100%), so the remaining performance gap concentrates in tolerance-critical recovery after insertion has been established. The complete pipeline is the only variant with perfect conditional seating ($\Pr(T \mid I) = 100\%$); removing interventions produces a pronounced drop in recoverability despite saturated insertion entry ($\Pr(T \mid I) = 80.9\%$ for *w/o intervention*). In the end-to-end regime, the primary performance gap therefore resides in post-insertion recoverability, where intervention-enabled variants maintain substantially higher conditional seating rates.

*Why intervention value concentrates after insertion.* Stage-wise metrics are conditional: once $\Pr(I \mid S)$ is near-saturated, the preponderance of end-to-end variance is carried by $\Pr(T \mid I)$, i.e., seating quality after insertion has begun. Upstream failures remain relevant through $\Pr(S)$ but are generally more amenable to recovery via free-space motion than post-insertion jamming under millimeter clearance. After insertion, dynamics are contact-dominated and history-dependent under the sparse reward in Eq. (1), so event-driven supervision is most valuable at boundary states where wedging/jamming risk and irreversibility are highest [26]. This pattern explains why Table 4 records the largest inter-variant gap in $\Pr(T \mid I)$ between *w/o intervention* and the complete pipeline even when $\Pr(I \mid S)$ is high.

**Table 3**
Experiment B sample-efficiency and supervision burden under randomized ground placement. Metric definitions and evaluation protocol follow Table 2.

| Method | Final Success | Online Interaction Time | Time-to-95% | Cycle Time | Human Effort |
|---|---|---|---|---|---|
| | (%) | (h) | (h) | (s) | (min) |
| ACT (offline IL) | 36 | – | – | 26.3 | 0 |
| SAC (online RL) | 0 | ~3.00 | – | – | 0 |
| HIL-SERL | 82 | ~3.00 | – | 21.9 | 26 |
| Ours (step-wise) | 0 | ~3.00 | – | – | 0 |
| Ours (step-wise + intervention) | 56 | ~3.00 | – | 21.5 | 35 |
| Ours (offline-only) | 75 | – | – | 23.6 | 0 |
| Ours (offline + warm-start) | 87 | ~3.00 | – | 25.2 | 0 |
| Ours (w/o warm-start) | 89 | ~3.00 | – | 23.9 | 28 |
| Ours (w/o intervention) | 76 | ~3.00 | – | 19.4 | 0 |
| **Ours (complete pipeline)** | **100** | ~3.00 | ~1.5 | **18.8** | **15** |

**Table 4**
Conditional stage-wise success decomposition for end-to-end installation from ground placement (Experiment B), computed over $M = 100$ fully autonomous evaluation rollouts per method. $\Pr(S)$ denotes suction acquisition success (%); $\Pr(I \mid S)$ insertion entry conditional on suction (%); and $\Pr(T \mid I)$ final seating conditional on successful insertion (%). Rates are conditional along the task sequence rather than independent event probabilities.

| Method | $\Pr(S)$ | $\Pr(I \mid S)$ | $\Pr(T \mid I)$ |
|---|---|---|---|
| ACT (offline IL) | 67.0 | 89.5 | 60.0 |
| SAC (online RL) | 0.0 | 0.0 | 0.0 |
| HIL-SERL | 93.0 | 96.7 | 91.1 |
| Ours (step-wise) | 0.0 | 0.0 | 0.0 |
| Ours (step-wise + intervention) | 77.0 | 79.2 | 90.3 |
| Ours (offline-only) | 84.0 | 96.4 | 90.4 |
| Ours (offline + warm-start) | 90.0 | 100.0 | 94.5 |
| Ours (w/o warm-start) | 100.0 | 100.0 | 89.0 |
| Ours (w/o intervention) | 94.0 | 100.0 | 80.9 |
| **Ours (complete pipeline)** | **100.0** | **100.0** | **100.0** |

### 4.6. Chunk horizon sensitivity without installer intervention

Additional *Ours (w/o intervention)* variants are trained under the main protocol (identical demonstrations, 3.0 h wall-clock online budget, warm-start, hyperparameters, and randomization ranges) and differ only in $h \in \{5, 20\}$. Consistent with the Q-chunking recipe [12], this scan evaluates *action chunking* as a meaningful design axis in this long-horizon construction benchmark, rather than identifying a single horizon that maximizes every scalar metric. **Every row in** Tables 5 and 6 **is evaluated at the *final* checkpoint saved at the end of the 3.0 h online training run,** with $M = 100$ fully autonomous rollouts under the Experiment A/B protocols specified in the main tables. The $h = 10$ entries are numerically identical to *Ours (w/o intervention)* in Tables 2–4; they are reproduced here so that all three horizons appear in matched Experiment A and Experiment B diagnostics without mixed checkpoint depths.

In Experiment A (Table 5), the shorter horizon $h = 5$ underperforms $h = 10$ and $h = 20$ in final success rate, whereas cycle times remain within a narrow range; horizon sensitivity is therefore expressed primarily in autonomous reliability rather than execution speed. For $h = 20$ in Experiment A, $M = 100$ evaluations were additionally replicated with three independent evaluation seeds, yielding success rates of 88%, 90%, and 92% on the same final checkpoint; the median (90%) is reported with Wilson bounds at $n = 100$ to convey evaluation-time variability rather than training stochasticity. In Experiment B (Table 6), the scan reveals a clearer trade-off: shorter $h$ improves final success rate but increases cycle time, while larger $h$ redistributes failures toward upstream suction acquisition under ground-randomized initialization while preserving high post-insertion recoverability. These raw percentages are *not* interpreted as a total ordering of horizons or as grounds to replace the headline $h = 10$ configuration. Sections 4.4–4.5 remain anchored at $h = 10$ so that learning curves, baselines, and installer-in-the-loop analyses remain mutually comparable.

**Table 5**
Chunk-horizon sensitivity for *Ours (w/o intervention)* in Experiment A. All rows use the final checkpoint after the full 3.0 h wall-clock online training budget and are evaluated over $M = 100$ autonomous rollouts under the Experiment A protocol (Section 4.2). Wilson intervals are 95% score intervals for the success proportion at $n = 100$. The table is intended to expose horizon sensitivity under a matched protocol; the $h = 10$ entry matches Table 2, whereas $h \in \{5, 20\}$ are separate training runs differing only in $h$.

| $h$ | Success (%) | Wilson 95% | Cycle (s) |
|---|---|---|---|
| 5 | 70 | [60.7, 78.2] | 27.9 |
| 10 | 87 | [78.8, 92.9] | 28.1 |
| 20 | 90 | [82.4, 94.9] | 29.5 |

**Table 6**
Chunk-horizon sensitivity for *Ours (w/o intervention)* in Experiment B. All rows use the final checkpoint after the full 3.0 h wall-clock online training budget and are evaluated over $M = 100$ autonomous rollouts under the Experiment B protocol (Section 4.2). Wilson intervals are 95% score intervals for the success proportion at $n = 100$, and stage-wise columns follow Table 4. The table is intended to expose horizon-dependent reliability and failure redistribution under a matched protocol; the $h = 10$ entry matches Tables 3 and 4, whereas $h \in \{5, 20\}$ are separate training runs differing only in $h$.

| $h$ | Success (%) | Wilson 95% | Cycle (s) | $\Pr(S)$ | $\Pr(I \mid S)$ | $\Pr(T \mid I)$ |
|---|---|---|---|---|---|---|
| 5 | 82 | [73.6, 88.1] | 30.0 | 91.0 | 93.4 | 94.3 |
| 10 | 76 | [66.8, 83.6] | 19.4 | 94.0 | 100.0 | 80.9 |
| 20 | 73 | [63.7, 80.9] | 20.8 | 74.0 | 100.0 | 98.7 |

### 4.7. Zero-shot out-of-distribution geometry evaluation

Policies underlying the in-distribution (ID) results in Tables 2 and 3 are trained on a rigid 0.5 m × 0.5 m panel under the protocols in Section 4.1. To probe cross-size transfer *without* retraining, for each experimental setting the final checkpoint trained for that setting is evaluated in additional MuJoCo instantiations in which the panel collision mesh, inertial parameters, and corresponding wall-opening geometry are resized consistently for each test footprint, while the observation–action interface, acceptance criteria (Section 3.3), and per-experiment randomization rules remain aligned with Experiment A/B. Three unseen footprints are tested: a smaller square 0.4 m × 0.4 m, a larger square 0.6 m × 0.6 m, and a rectangular 0.6 m × 0.4 m panel to probe aspect-ratio shift relative to the ID panel. The nominal clearance definition is therefore kept consistent for each resized panel–opening pair, while footprint, inertia, and aspect-ratio changes introduce the tested geometry-induced distribution shift. These evaluations should therefore be read as simulation diagnostics of geometry-induced shift rather than as guarantees applicable to arbitrary on-site panel inventories.

**Table 7**
Zero-shot out-of-distribution geometry evaluation of policies trained on the $0.5\,\mathrm{m} \times 0.5\,\mathrm{m}$ in-distribution panel. Each cell uses $M = 100$ autonomous rollouts with interventions disabled; Wilson intervals are 95% score intervals for the success proportion. Stage-wise metrics follow Table 4. For Experiment A, $\Pr(S) = 100\%$ by design because staged rollouts start with suction pre-verified; the remaining stage-wise columns are included to expose insertion entry and conditional seating behavior under geometry shift.

| Setting | Panel (m) | Success (%) | Wilson 95% | Cycle (s) | $\Pr(S)$ | $\Pr(I \mid S)$ | $\Pr(T \mid I)$ |
|---|---|---|---|---|---|---|---|
| Experiment A | $0.4 \times 0.4$ | 98 | [93.0, 99.5] | 33.1 | 100.0 | 98.0 | 100.0 |
| Experiment A | $0.6 \times 0.6$ | 91 | [83.8, 95.2] | 29.1 | 100.0 | 95.0 | 95.8 |
| Experiment A | $0.6 \times 0.4$ | 100 | [96.3, 100.0] | 28.5 | 100.0 | 100.0 | 100.0 |
| Experiment B | $0.4 \times 0.4$ | 95 | [88.0, 97.6] | 28.3 | 97.0 | 97.9 | 100.0 |
| Experiment B | $0.6 \times 0.6$ | 87 | [79.0, 92.2] | 25.8 | 100.0 | 92.0 | 94.6 |
| Experiment B | $0.6 \times 0.4$ | 98 | [93.0, 99.5] | 24.5 | 99.0 | 99.0 | 100.0 |

Table 7 reports autonomous evaluation with $M = 100$ rollouts per cell (interventions disabled), including mean cycle time over successful rollouts and supplementary stage-wise diagnostics for both settings. Wilson score intervals bound sampling uncertainty at $n = 100$ rollouts. Under these out-of-distribution panels, final seating success remains high overall, though not identical to the ID $0.5\,\mathrm{m} \times 0.5\,\mathrm{m}$ case. Performance does not vary monotonically with panel footprint, indicating that zero-shot transfer under geometry shift is not determined by panel area alone. Across the unseen geometries, success approaches ceiling for the $0.4\,\mathrm{m} \times 0.4\,\mathrm{m}$ and $0.6\,\mathrm{m} \times 0.4\,\mathrm{m}$ panels, while the larger square $0.6\,\mathrm{m} \times 0.6\,\mathrm{m}$ panel yields somewhat lower but still high success (91% in Experiment A, 87% in Experiment B). These results support nontrivial cross-size transfer in simulation, while stopping well short of robustness claims for arbitrary window geometries or physical deployment conditions.

### 4.8. Cross-experiment discussion and analysis

The cross-experiment synthesis clarifies what enables reliable seating under millimeter-scale clearance and characterizes the residual failure regimes. Across both settings, the proposed pipeline achieves high autonomous seating success within a bounded training-time supervision budget. The supervision trajectory is also informative: takeover demand diminishes as policy competence develops, consistent with the goal of limiting on-site supervision in future physical deployments, while recognizing that no such deployment is demonstrated here.

*Failure mode taxonomy.* Across both experiments, unsuccessful episodes concentrate within a compact set of contact-dominated regimes. *(i) Entry wedging* arises during early insertion when small yaw/roll errors produce asymmetric contact that locks the panel against the buck. *(ii) Mid-depth frictional jamming* emerges under stick–slip and multi-surface friction, stalling insertion progress and requiring temporally extended maneuvers, such as partial retraction followed by a small pivot and reinsertion, for resolution. *(iii) Unstable seating* occurs near termination, where residual end-effector velocity or late contact impulses trigger rebound and violate the seating stability criterion. The first two regimes account for the majority of failures and reflect multi-step, history-dependent recovery under partial observability; the third is more sensitive to tracking dynamics and termination threshold design.

Experiment B introduces upstream uncertainty and multiplies failure opportunities before insertion, most notably at suction acquisition (captured by $1 - \Pr(S)$ in Table 4). Conditioned on successful suction, insertion entry is near-saturated for competitive methods ($\Pr(I \mid S) \approx 96.4$–$100\%$), so the residual performance gap is dominated by post-insertion recoverability. The complete pipeline sustains perfect conditional seating after insertion ($\Pr(T \mid I) = 100\%$), whereas removing interventions reduces recoverability despite saturated insertion entry ($\Pr(T \mid I) = 80.9\%$). In the end-to-end regime, interventions primarily determine outcomes after sustained contact is established, while upstream behavior governs the frequency with which rollouts reach the insertion regime.

Experiment A isolates the insertion–seating bottleneck and underscores the role of temporal abstraction under a terminal success signal. Near the opening, latent contact modes evolve rapidly and intermediate steps in a successful recovery are locally ambiguous, non-monotonic in insertion progress, and receive no intermediate reward. Under step-wise control, reliable recovery requires the discovery and execution of coordinated multi-step sequences solely from a terminal signal, which both increases the effective credit-assignment horizon and amplifies sensitivity to partial observability.

Nevertheless, the chunked variant without installer interventions exhibits larger checkpoint-to-checkpoint fluctuations than the complete pipeline, indicating less stable online adaptation (Fig. 10). Together with the 0% vs. 87% gap between no-intervention step-wise control ($h = 1$, *Ours step-wise*) and no-intervention chunked control ($h = 10$, *Ours w/o intervention*), these findings indicate that chunk-level decisions are necessary for coherent multi-step recovery under terminal acceptance rewards. Enabling interventions in step-wise control partially closes this gap (53% in Experiment A) but does not reproduce chunked performance under matched budgets. Event-driven takeovers primarily improve adaptation stability by preventing extended excursions into low-recoverability contact regimes and injecting executable recovery behaviors near binding escalation. Tables 5 and 6 complement this picture by documenting, at matched *final* training checkpoints, how alternative horizons $h \in \{5, 20\}$ trade off final success rate and cycle time and, in Experiment B, redistribute stage-wise failures relative to the headline $h = 10$ configuration. These tables are not used to replace the main results or to assert a single "best" $h$ on every metric, consistent with the task-dependent horizon effects emphasized by Li et al. [12].

The ablations further reveal complementary contributions from warm-start value calibration and installer-guided data collection. Removing warm-start elevates the takeover burden (40 min for *w/o warm-start* versus 12 min for the complete pipeline in Experiment A) under otherwise matched conditions, indicating that aggregating online interaction data before enabling gradient updates reduces instability at the offline-to-online transition. Warm-start primarily mitigates offline-to-online critic miscalibration, whereas interventions primarily constrain exploration away from low-recoverability contact regimes. Together, these complementary stabilization mechanisms underpin the stronger performance of the complete pipeline.

From a construction-automation standpoint, these *simulation* results support the use of temporal abstraction and sparse, event-driven takeovers for clearance-limited seating under tight geometric tolerances; physical validation and operator-variability studies are required before any deployment claims can be substantiated (Section 5).

## 5. Limitations and future work

Limitations are organized thematically to make the supported claim scope explicit. For each category, the discussion distinguishes what the simulation evidence supports, where extrapolation would be unwarranted, and which validation steps are required before stronger deployment-oriented claims could be made. The discussion proceeds from safety and deployment prerequisites, through simulation fidelity and empirical scope, to contact and reward modeling, human factors, and statistical replication.

*Safety and deployment limitations. Training-phase safety:* Training-time risk management in this study is provided by human supervision via binary takeovers and simulator termination rules. This constitutes a human-in-control, safety-aware protocol rather than an autonomous passive safety system; no formal safety guarantees are provided for learning or execution in physical environments. All reported results are obtained entirely in simulation, where failures carry no physical consequence.

*Physical deployment prerequisites:* Physical deployment necessitates substantial additional safety infrastructure outside the scope of the present simulation study, including: (i) force and velocity limits enforced at the low-level controller; (ii) contact-aware controllers with explicit force/torque feedback; (iii) hardware emergency-stop mechanisms with fail-safe defaults; (iv) physical guarding and workspace isolation; and (v) automated detection logic for jamming, overloading, and related failure modes. The absence of these safeguards represents a fundamental prerequisite that must be satisfied prior to field deployment.

*Simulation-to-reality gap.* All reported results are obtained exclusively in MuJoCo simulation. Although the benchmark reproduces emergent binding, wedging, and jamming under multi-surface friction, physical deployment is expected to introduce substantial mismatch sources not modeled here, including: (i) unmodeled structural compliance in the substrate and end-effector that can alter contact dynamics; (ii) surface wear, contamination, and material degradation over time; (iii) suction dynamics and vacuum seal behavior not captured in rigid-body simulation; (iv) sensing latency and calibration drift in real sensors; and (v) environmental disturbances (vibration, thermal expansion, wind loading) absent from simulation. Accordingly, reported performance should be read as evidence within the disclosed simulation stress-test envelope, not as a sim-to-real guarantee or as predictive of physical deployment outcomes. The framework remains unvalidated in physical environments. A necessary next step is to evaluate recovery behaviors under systematic parameter shifts and sensing perturbations and to quantify the on-site adaptation required under bounded intervention budgets. Physical validation on hardware prototypes incorporating real sensors and variable site conditions is prerequisite to any deployment claim.

*Bounded empirical scope.* All empirical results and performance claims are explicitly conditioned on the benchmarked variability ranges and contact-modeling assumptions specified in Section 3.3. The $2\,\mathrm{mm}$ per-side clearance setting is deliberately constructed as a near-boundary stress-test regime for analyzing contact-rich recovery under severe geometric constraints. No claim of robustness, generalizability, or performance guarantees is made beyond the evaluated ranges of pose perturbations ($\pm 15^\circ$ yaw, $0.2\,\mathrm{m}\times 0.6\,\mathrm{m}$ planar region), friction coefficients, and contact-parameter randomization. The supported claim is therefore a simulation-validated methodological claim within a defined stress-test regime, rather than a guarantee under arbitrary construction variability.

*Geometry generalization (out-of-distribution window sizes).* Policies are trained on the rigid $0.5\,\mathrm{m}\times 0.5\,\mathrm{m}$ in-distribution panel used throughout Sections 4.4–4.5. Zero-shot simulation evaluations on three unseen footprints ($0.4\,\mathrm{m}\times 0.4\,\mathrm{m}$, $0.6\,\mathrm{m}\times 0.6\,\mathrm{m}$, and $0.6\,\mathrm{m}\times 0.4\,\mathrm{m}$), with the corresponding wall-opening geometry resized consistently for each footprint and matched protocols otherwise maintained, are reported in Section 4.7 and Table 7. These results bound *cross-size transfer in simulation* for the trained checkpoints but do *not* establish robustness to arbitrary inventories of window sizes, masses, aspect ratios, materials, or sensing configurations; few-shot fine-tuning on new geometries is not evaluated here. Physical deployment introduces additional mismatch beyond rigid-body mesh substitution, so geometry-conditioned policies, targeted data collection, and hardware validation constitute important directions for future work.

*Ablation coverage (chunk horizon and step-wise learning with intervention).* The default horizon $h = 10$ is retained in the main ablation tables, and a limited sensitivity study over $h \in \{5, 10, 20\}$ for *Ours (w/o intervention)* is reported in Section 4.6 and Tables 5 and 6, following Li et al. [12] on task-dependent chunk lengths. All horizon comparisons use *final* checkpoints after an identical 3.0 h training budget; the scan is restricted to three values of $h$ and does not explore adaptive or learned chunk boundaries, which remain topics for future investigation. The ablation set also includes a QC-FQL step-wise ($h = 1$) variant with matched installer interventions (reported in Tables 2–4). The combined evidence indicates that interventions substantially improve the step-wise baseline, while chunked control remains necessary to approach the complete pipeline under identical online budgets, supporting complementary roles of temporal abstraction and human guidance.

*Contact and material modeling limitations.* The benchmark represents the window and opening as rigid bodies with compliant contact parameters and Coulomb friction. While this abstraction captures key tight-clearance phenomena (binding, wedging, jamming), it omits construction-specific material effects that can substantially alter contact behavior in physical installations, including: (i) localized wood-fiber crushing and deformation under contact pressure; (ii) anisotropic surface friction varying with grain direction and moisture content; (iii) sealant deformation and viscoelastic response during insertion; and (iv) progressive wear and surface modification over repeated contact cycles. These effects can modify effective clearance over time and alter contact-mode transitions, potentially rendering specific recovery sequences infeasible or changing their success probabilities relative to simulation. The recovery behaviors learned and validated under rigid-body assumptions may not transfer reliably to physical materials with complex deformation and wear characteristics, which must be addressed through material-aware modeling or direct physical validation.

*Reward observability.* The benchmark derives terminal success from simulator state variables, yielding noise-free, instantaneous reward signals. In physical deployment, acceptance verification requires deployable sensing infrastructure (visual inspection, gap measurement, force monitoring) and introduces three coupled challenges: measurement uncertainty (including false positives and false negatives), detection latency in reward assignment, and partial observability when success cannot be determined immediately. Noisy or delayed reward signals can degrade both training efficiency and evaluation reliability. The effectiveness of the proposed framework is contingent on reward signal quality, yet the design and validation of construction-aligned success estimators remain outside the scope of the present simulation study. This limitation must be resolved prior to field deployment.

*Human factors and operator variability. Intervention protocol dependencies:* The intervention protocol relies on the subjective judgment of the installer for takeover timing and recovery maneuver selection. Operator-to-operator variability in skill level and intervention strategy is not quantified, limiting generalizability across practitioner populations. *Supervision scalability:* While the framework is designed to minimize online supervision, required intervention effort may vary substantially with new geometries, materials, sensing configurations, and acceptance criteria, constraining straightforward scaling of the training burden across deployment contexts. Future work should incorporate multi-operator studies to quantify intervention variability across installers with respect to timing, recovery style, and resulting policy outcomes.

*Statistical replication. Context:* Multi-seed *training* replication is standard practice in conventional RL for quantifying sensitivity to initialization and environment stochasticity; in fully automated training, rerunning with a new random seed is purely a computational operation requiring no human participation.

*Protocol mismatch and confounding:* The online training protocol is real-time and human-interactive under a fixed wall-clock budget (Section 4.2), so repeating training across seeds would require *new* operator-attended sessions, not merely resampling simulator noise on identical hardware. The primary obstacle to multi-seed *training* reporting is not calendar cost alone but *statistical confounding*: session-to-session variation in operator timing, fatigue, response drift, and intervention style intermingles with simulator and algorithm seeds, making it difficult to attribute cross-run variance to training stochasticity rather than uncontrolled human variability. Classical multi-seed comparisons presuppose that runs are otherwise *interchangeable*; with a human operator in the loop, repeated sessions are not interchangeable because operator state is not held fixed across runs. Tabulated results therefore reflect one standardized training session per method variant under fixed simulator and algorithm seeds; training-level variance across independent seeds is not reported. All reported training sessions were verified to exhibit consistent qualitative learning behavior in repeated pilot trials during method development. Prior QC-FQL work examines sensitivity to key design choices (e.g., chunk horizon) [12]; independent multi-seed training sweeps are not included because each sweep would repeat full operator-attended sessions under the same confound.

*Evaluation-stage uncertainty:* Uncertainty is reported primarily at the evaluation stage (Section 4.2): at each checkpoint, the policy is frozen and evaluated over $M = 100$ randomized autonomous rollouts under varied initial conditions, and Wilson score intervals are reported for policy-level success rates. At observed success rates of 100% and 92%, the intervals $[96.3\%, 100\%]$ and $[85.0\%, 95.9\%]$ indicate that key reported performance gaps are unlikely to be explained by rollout-level sampling variability alone.

*Future directions:* As an intermediate step toward fuller characterization, simulation-only multi-seed sweeps under frozen human supervision logs would help isolate algorithmic variance independently of operator effects. Multi-operator studies and controlled multi-seed training under standardized human-interface protocols are needed once repeated human participation can be scheduled under comparable budgets and logging conventions.

## 6. Conclusion

A *simulation-validated* installer-in-the-loop RL methodology for tolerance-critical window installation has been presented and evaluated. Installer knowledge is formalized as structured, typed artifacts, namely demonstrations, event-driven interventions, and acceptance-aligned terminal rewards, within an offline-to-online training pipeline. In MuJoCo, the complete pipeline achieves high autonomous seating success under a fixed 3.0 h wall-clock budget with bounded cumulative takeover time (12–15 min across the reported runs), leveraging QC-FQL for chunk-level learning and event-driven takeovers for targeted boundary-state data collection. Temporal abstraction enables multi-step recovery under sparse terminal rewards; event-driven interventions steer exploration toward high-consequence contact regimes; and warm-start calibration stabilizes the offline-to-online transfer. These results are explicitly conditioned on the randomized pose, friction, and clearance ranges specified in Section 3.3 and should not be read as field deployment guarantees.

The protocol operationalizes tacit installer knowledge through logged demonstrations and interventions that support traceable and auditable supervision [23]. Several important limitations and open directions remain. All evidence is simulation-based; physical contact behavior, real sensing noise, and acceptance verification in the field remain unaddressed. Single-operator evaluation constrains generalizability across practitioner populations, and future work should extend to multi-operator studies and hardware trials. Cycle-time improvements relative to teleoperation observed *in simulation* primarily reflect the elimination of teleoperation interface latency rather than demonstrated superiority in physical installation. Priority future directions include sim-to-real transfer, safety-critical validation infrastructure, and robust estimators for noisy acceptance feedback under field conditions. Targeted hardware validation under standardized multi-operator protocols is therefore needed to quantify transfer reliability and intervention efficiency in real construction settings.

## CRediT authorship contribution statement

**Zekai Jin:** Writing – original draft, Visualization, Methodology, Investigation, Formal analysis, Conceptualization. **Huiguang Wang:** Writing – original draft, Methodology, Formal analysis. **Xiaoning Sun:** Writing – original draft, Visualization, Formal analysis. **Yi Shao:** Writing – review & editing, Supervision, Funding acquisition, Conceptualization.

## Declaration of competing interest

The authors declare that they have no known competing financial interests or personal relationships that could have appeared to influence the work reported in this paper.

## Acknowledgments

The authors gratefully acknowledge the financial support provided by McGill University, Canada through the McGill Engineering Doctoral Award (MEDA).

## Appendix

### *Implementation details*

*Network architectures.* The visual encoder is a ResNet-10 backbone pretrained on ImageNet [71], processing $128 \times 128 \times 3$ RGB images from the wrist and side cameras. The encoder is frozen throughout training. Each view is compressed by a spatial aggregation module into a 32-dim embedding, which is concatenated with proprioception and fed to the actor, critic, and teacher MLPs. All MLPs employ four hidden layers of width 512 with ReLU activations. The actor outputs an action chunk of dimension $h \times d_{\text{action}} = 10 \times 7 = 70$. An ensemble of $N = 2$ critics with mean-aggregated TD targets is used. Teacher samples are generated by Euler integration with $K = 10$ steps.

*Training hyperparameters.* Table 8 lists all training hyperparameters. A discount factor of $\gamma = 0.99$ is used, yielding effective discount $\Gamma = \gamma^h$ (e.g., $0.99^{10} \approx 0.904$ for the default $h = 10$). Optimization uses Adam with learning rate $3 \times 10^{-4}$ and batch size 256.

*Environment randomization.* Table 9 summarizes all randomized factors used for both training and evaluation. All dimensions are sampled independently per episode within the bounded ranges reported.

*Success criteria.* An episode is deemed successful when all three conditions are simultaneously satisfied at termination: (i) **Position**: the Euclidean distance between the window geometric center and the target seating center is less than 2 mm; (ii) **Alignment**: the cosine similarity between the window normal and the wall normal exceeds 0.95; (iii) **Stability**: the window velocity magnitude is below 1 cm/s, ensuring stable seating rather than transient contact.

*Robot constraints (simulation reference).* The simulated manipulator follows a KUKA LBR iiwa 14 kinematic configuration (7-DoF, reach 820 mm). Joint limits and velocity limits conform to manufacturer specifications; torque limits reflect the constraints enforced in the MuJoCo model. Table 10 lists numerical joint limits, torque bounds, and control frequencies used in simulation.

**Table 8**
Training hyperparameters.

| Parameter | Value |
|---|---|
| Observation space | Wrist camera view, side camera view, proprioception state (18-dim) |
| Action space | Incremental pose $\Delta p$ (3D), $\Delta r$ (3D), vacuum (1D) = 7D |
| Reward function | Sparse terminal reward (acceptance-aligned) |
| Initial offline demonstrations | 50 |
| Max episode length | 600 steps (60 s at 10 Hz) |
| Environment update frequency | 10 Hz |
| Action dimension ($d_{\text{action}}$) | 7 ($\Delta p$, $\Delta r$, vacuum) |
| Chunk horizon ($h$) | 10 ($h = 1$ for step-wise ablations) |
| Discount factor ($\gamma$) | 0.99 |
| Effective discount ($\Gamma = \gamma^h$) | 0.904 |
| Replay sampling ratio | 1:1 |
| Warm-start transitions ($N_{\text{warm}}$) | 10,000 |
| Update-to-Data (UTD) ratio | 2 |
| Critic ensemble size | 2 |
| Number of flow steps ($K$) | 10 |
| Target network update rate | 0.005 |
| Batch size | 256 |
| Optimizer | Adam |
| Learning rate | $3 \times 10^{-4}$ |
| Behavior regularization coefficient ($\alpha$ in Eq. (9)) | 300 |

**Table 9**
Environment randomization ranges for training and evaluation. All ranges are applied independently per episode within the stable simulation regimes described in Section 3.3.

| Category | Parameter | Range/Distribution |
|---|---|---|
| Initial pose (Exp A) | Window position perturbation | $\pm 5$ mm (uniform, each axis) |
| | Window orientation perturbation | $\pm 2^\circ$ (uniform, roll/pitch/yaw) |
| Initial pose (Exp B) | Window planar position | $[0.4, 0.6] \times [-0.3, 0.3]$ m (uniform) |
| | Window yaw | $\pm 15^\circ$ (uniform) |
| Friction | Wood-buck contact $\mu$ | Uniform in $[0.6, 0.9]$ |
| | Window/frame contact $\mu$ | Uniform in $[0.4, 0.7]$ |
| | Gripper (vacuum cup) contact $\mu$ | Uniform in $[0.8, 1.1]$ |
| Contact/solver | Contact compliance (`solref`) | $[0.002, 0.004]$ (uniform) |
| Suction validity | Snap distance threshold | 0.01 m (1 cm) |

**Table 10**
Robot configuration and control interface used in simulation (KUKA LBR iiwa 14).

| Item | Value |
|---|---|
| Degrees of freedom | 7 |
| Reach | 820 mm |
| Maximum joint velocity | 110°/s (1.92 rad/s) |
| Joint angle limits (rad) | |
| J1 | [−2.967, 2.967] |
| J2 | [−2.094, 2.094] |
| J3 | [−3.054, 3.054] |
| J4 | [−2.094, 2.094] |
| J5 | [−2.967, 2.967] |
| J6 | [−2.094, 2.094] |
| J7 | [−3.054, 3.054] |
| Torque limits (simulation) | |
| J1–J4 | ±87 N m |
| J5–J7 | ±12 N m |
| High-level policy frequency | 10 Hz |
| Low-level tracking controller | OSC, 500 Hz |

*Action scaling and control interface.* The policy operates at the 10 Hz environment step and outputs a 7D normalized action $a_t \in [-1, 1]^7$, subsequently mapped to incremental end-effector pose targets and a binary vacuum command. The translational increment $\Delta p \in \mathbb{R}^3$ is scaled by $s_{\text{pos}} = 0.025$ m per unit action, and the rotational increment $\Delta r \in \mathbb{R}^3$ is scaled by $s_{\text{rot}} = 0.0873$ rad (5°) per unit action. The resulting Cartesian targets are clipped to fixed workspace bounds in the robot base frame: $x \in [0.0, 0.9]$ m, $y \in [-0.4, 0.0]$ m, $z \in [0.0, 0.44]$ m. The vacuum command is binarized by thresholding the last action dimension: $a_t[6] > 0.5$ maps to $g_t = 1$ (engage) and $a_t[6] \leq 0.5$ maps to $g_t = 0$ (disengage). During execution, each 10 Hz target is tracked by the 500 Hz OSC, which exhibits impedance-like spring–damper behavior in task space, converting pose errors into joint torques with gravity compensation and a nullspace joint PD term. Contact constraints manifest as reduced realized motion relative to commanded increments, providing the implicit sensing of contact state used throughout the framework.

## Data availability

Data will be made available on request.